\documentclass[a4paper]{article}

\usepackage[margin=1in]{geometry}
\usepackage{hyperref}
\usepackage{amsmath}
\usepackage{amsthm}
\usepackage{amsfonts,amssymb}
\usepackage{indentfirst}
\usepackage{enumerate}%9.4
\usepackage{color}
\usepackage{graphicx}
\usepackage{subcaption}
\usepackage{multirow}
\usepackage{geometry}
\usepackage{authblk}

\begin{document}

\title{Binary structured physics-informed neural networks for solving equations with rapidly changing solutions}

\author[1]{Yanzhi Liu}
\author[1]{Ruifan Wu}
\author[1]{Ying Jiang \thanks{Corresponding author: jiangy32@mail.sysu.edu.cn}}
\affil[1]{School of Computer Science and
Engineering, Guangdong Province Key Laboratory of Computational Science, Guangzhou, 510006, People’s Republic of China}
\maketitle

% abstract
\begin{abstract}
Physics-informed neural networks (PINNs), rooted in deep learning, have emerged as a promising approach for solving partial differential equations (PDEs).
By embedding the physical information described by PDEs into feedforward neural networks, PINNs are trained as surrogate models to approximate solutions without the need for label data.
Nevertheless, even though PINNs have shown remarkable performance, they can face difficulties, especially when dealing with equations featuring rapidly changing solutions. These difficulties encompass slow convergence, susceptibility to becoming trapped in local minima, and reduced solution accuracy.
To address these issues, we propose a binary structured physics-informed neural network (BsPINN) framework, which employs binary structured neural network (BsNN) as the neural network component.
By leveraging a binary structure that reduces inter-neuron connections compared to fully connected neural networks, BsPINNs excel in capturing the local features of solutions more effectively and efficiently. These features are particularly crucial for learning the rapidly changing in the nature of solutions.
In a series of numerical experiments solving Burgers equation, Euler equation, Helmholtz equation, and high-dimension Poisson equation, BsPINNs exhibit superior convergence speed and heightened accuracy compared to PINNs. From these experiments, we discover that BsPINNs resolve the issues caused by increased hidden layers in PINNs resulting in over-smoothing, and prevent the decline in accuracy due to non-smoothness of PDEs solutions.
\end{abstract}

% introduction
\section{Introduction}
Partial differential equations (PDEs) are essential mathematical tools that describe physical phenomena such as electricity \cite{zhang2022dynamic}, heat \cite{avalos2013rational}, and fluid flow \cite{arqub2018numerical}. They are crucial for understanding and solving natural phenomena and engineering problems.
However, acquiring analytical solutions for PDEs can be exceedingly challenging, especially when dealing with highly nonlinear, unconventional boundary conditions, or intricate geometric shapes \cite{de2022error}.
Therefore, computing numerical approximations for PDEs quickly and accurately holds significant importance.
In the field of numerical methods for solving PDEs, there are many widely used methods, such as the finite element method \cite{ern2004theory}, the finite volume method \cite{eymard2000finite}, the spectral method \cite{trefethen2000spectral}, and the finite difference method \cite{chapra2010numerical}, often referred to as traditional numerical methods.
However, to achieve higher accuracy, traditional numerical methods often require sufficiently fine grids of solution domains, which can result in significant computational costs \cite{Washington_Buja_Craig_2009}.
Additionally, when addressing PDEs characterized by state or parameter spaces with high dimensions, traditional numerical methods necessitate the subdivision of high-dimensional grids. Consequently, as the dimensionality increases, the computational cost will increase exponentially \cite{highorder}, occasionally rendering the computations infeasible. This challenging computational issue is commonly referred to as the ``curse of dimensionality'' \cite{Darbon_Osher_2016}.

To address the aforementioned challenges related to conventional numerical methods, machine learning techniques have been employed for solving PDEs since the 1990s \cite{1994Neural}. In particular, neural networks such as fully connected neural networks (FNNs), recurrent neural networks (RNNs), and residual neural networks (ResNets) are widely used in solving PDEs \cite{ling2016reynolds,parish2016paradigm,rico1994continuous,vlachas2018data}. These neural network-based methods require a substantial amount of accurately resolved or experimentally measured data for training.
However, the cost of data acquisition in the fields of physics or biology modeling is prohibitively high. Consequently, we are often compelled to draw conclusions and make decisions in scenarios where only a limited amount of data or no data is available \cite{Raissi_Perdikaris_Karniadakis_2018}.
With the explosive growth of scientific machine learning \cite{ScientificML} and automatic differentiation techniques \cite{Baydin_Pearlmutter_Radul_Siskind_2015} in recent years, a novel deep learning-based approach called physics-informed neural networks (PINNs) \cite{Raissi_Perdikaris_Karniadakis_2018} has been proposed to overcome the reliance on training data by leveraging prior knowledge, such as physical laws.
PINNs perform well in solving a wide range of PDEs, including the Navier-Stokes equations for incompressible flow, the Korteweg-de Vries equation involving third-order derivatives, and the Schr\"{o}dinger equation with periodic boundary conditions.
Building upon the PINNs, numerous variations have been subsequently proposed. For instance, spline-PINN \cite{wandel2022spline}, which combines Hermite splines with convolutional neural networks, and NSFnets \cite{jin2021nsfnets}, employed for solving the Navier-Stokes equations, among others.

Nevertheless, PINNs encounter challenges related to convergence and accuracy degradation, especially when dealing with solutions that exhibit rapid changes at specific locations \cite{burden2015numerical,Wang_Teng_Perdikaris}. To address this issue, researchers have introduced numerous enhancement techniques building upon PINNs to accelerate the training of PINNs while enhancing their accuracy \cite{hu2022discontinuity, jagtap2020adaptive, lu2021deepxde, mao2020physics, mcclenny2020self, wang2022and, zeng2022adaptive}.

In \cite{hu2022discontinuity}, the problem of fitting $d$-dimensional discontinuous functions is transformed into the problem of fitting $(d+1)$-dimensional continuous functions, allowing neural networks to achieve a good fit. However, knowing the precise locations of the discontinuities is required, and in practice, this information is often unknown.

In \cite{mao2020physics}, an approach is introduced to enhance the performance of PINNs in solving the Euler equations that model high-speed aerodynamic flows. This enhancement is achieved by employing clustered training points instead of the commonly used randomly distributed training points. The clustered training points are created by densely placing training points near the locations where the solution changes rapidly. However, determining the precise locations of numerical singularities can still be a challenge, especially when these regions have complex distributions.

To solve 'stiff' PDEs, which have solutions containing sharp spatial transitions or fast temporal evolution, an adaptive algorithm is proposed in \cite{mcclenny2020self}. This algorithm assigns a trainable weight parameter to each training point. During the training process, the weights of training points in challenging-to-solve regions increase, directing PINNs to pay more attention to these difficult areas.

To address the issues of slow convergence or convergence difficulties for trainning PINNs, an optimization algorithm with adaptive weights is proposed \cite{wang2022and}. In this algorithm, the loss weights corresponding to the partial differential equations and boundary conditions are computed using the eigenvalues of the neural tangent kernel matrix at each iteration.

In \cite{jagtap2020adaptive, lu2021deepxde, zeng2022adaptive}, adaptive techniques including adaptive loss functions, adaptive activation functions, and adaptive sampling are employed to accelerate the training of PINNs while enhancing their accuracy.

The aforementioned techniques for enhancing the performance of PINNs often require prior information about equation solutions, or an increase in the number of parameters. The adaptive techniques often result in higher memory consumption and additional training parameters for PINNs.

In order to address the challenges associated with using PINNs to solve PDEs with rapidly changing solutions, we propose a framework called binary structured physics-informed neural network (BsPINN).
In contrast to PINNs, BsPINNs can address equations with rapidly changing solutions using fewer parameters and without requiring knowledge about the locations of challenging-to-solve regions
To evaluate the performance of BsPINNs, we employ them to tackle a wide range of PDEs, including Burgers equation, Euler equation, Helmholtz equation, and high-dimension Poisson equation, and conduct comparative analyses with PINNs.

Our experimental findings reveal that, in contrast to PINNs, BsPINNs exhibit notable advantages, such as enhanced convergence speed and superior solution accuracy, particularly in scenarios involving PDEs with rapidly-changing solutions. Additionally, based on our experimental findings, we observe that BsPINNs overcome the issue of ``over-smoothing" encountered by PINNs (refer to \cite{2020graph}).

Moreover, BsPINNs also address the problem of low accuracy in solving PDEs with discontinuous solutions, a challenge faced by PINNs. When neural networks are used to solve PDEs with discontinuous solutions, the training points near the discontinuity region often exhibit significant training losses \cite{mao2020physics}, exerting a substantial impact on the accuracy of the solution. These points are commonly referred to as transition points \cite{shen2014improvement}. In practice, when sampling piecewise continuous functions, it is observed that the proportion of transition points among the sampled points is typically very low, and this proportion tends to decrease as the number of sampled points increases. In other words, the distribution of training points exhibits a long-tailed phenomenon and transition points belong to the ``tail category" of the training points, making them more likely to be neglected by fully connected neural networks during the training process \cite{resnick2007heavy}. Consequently, this leads to low approximation accuracy of PINNs when dealing with discontinuous solution functions. Conversely, the proposed BsPINNs in this paper, due to its emphasis on local variations in functions, achieve higher accuracy when solving PDEs with discontinuous solutions.

The rest of the paper is organized as follows. In Section \ref{s2}, we provide a brief review of the mechanism employed by PINNs for solving PDEs. In Section \ref{s3}, we introduce the motivations and inspirations behind the binary network structure in BsPINNs and present its mathematical representation, followed by an example of function approximation to confirm its accuracy. Section \ref{s4} is dedicated to utilizing BsPINNs to solve a wide range of PDEs and assessing their performance by comparing them with the corresponding PINNs in terms of both convergence speed and solution accuracy. Finally, some concluding remarks and future work are presented in Section \ref{s5}.

\section{Physics-Informed Neural Networks}\label{s2}
In this section, we review the mechanism of PINNs used to tackle PDEs. Further elaboration can be found in \cite{Raissi_Perdikaris_Karniadakis_2018}. Consider the following initial-boundary value problem
\begin{flalign}
    \label{eqn}
    &\ \ \ \ \ \ \ \ L(u(x,t))=0 ,\quad x \in \Omega,\ t \in (0,T],& \\
    \label{boundary}
    &\ \ \ \ \ \ \ \ B(u(x,t))=\psi(x,t) ,\quad x \in \partial\Omega,\ t \in (0,T],&\\
    \label{initial}
    &\ \ \ \ \ \ \ \ u(x,t)=\varphi(x) ,\quad x \in \bar{\Omega},\ t=0.&
\end{flalign}
In this context, \eqref{eqn}, \eqref{boundary}, and \eqref{initial} correspond to the governing equations, boundary conditions, and initial conditions, respectively.
The function $u$ represents the solution of the PDE over the domain $\bar{\Omega} \times [0, T]$. To construct the loss function for PINNs, we introduce the following notations:
\begin{flalign}
    \label{eqn_r}
    &\ \ \ \ \ \ \ \ R_L(u(x,t)):=L(u(x,t))-0 ,\quad x \in \Omega,\ t \in (0,T], &\\
    \label{boundary_r}
    &\ \ \ \ \ \ \ \ R_B(u(x,t)):=B(u(x,t))-\psi(x,t) ,\quad x \in \partial\Omega,\ t \in (0,T],&\\
    \label{initial_r}
    &\ \ \ \ \ \ \ \ R_I(u(x,t)):=u(x,t)-\varphi(x) ,\quad x \in \bar{\Omega},\ t=0.&
\end{flalign}
These three quantities can be interpreted as the residuals originating from the governing equations, boundary conditions, and initial conditions.

Let $u_{\theta}$ represent the output obtained by using PINNs parameterized by $\theta$, which encompasses the network's weights and biases. By replacing $u$ in \eqref{eqn_r}, \eqref{boundary_r}, and \eqref{initial_r} with $u_{\theta}$, we derive the foundational principle for constructing the loss function utilized during the training process of PINNs. These are denoted as $R_L\left(u_{\theta}\left(x,t\right)\right)$, $R_B\left(u_{\theta}\left(x,t\right)\right)$, and $R_I\left(u_{\theta}\left(x,t\right)\right)$, respectively. Here, the following partial derivatives
\begin{flalign*}
    &\ \ \ \ \ \ \ \ \frac{\partial u_{\theta}(x,t)}{\partial x},\ \frac{\partial u_{\theta}(x,t)}{\partial t},\ \frac{\partial^2 u_{\theta}(x,t)}{\partial x^2},\ \frac{\partial^2 u_{\theta}(x,t)}{\partial t^2},\ldots&
\end{flalign*}
are calculated by utilizing the approach of automatic differentiation \cite{Baydin_Pearlmutter_Radul_Siskind_2015}.
In order to satisfy the PDEs, the ideal $u_{\theta}$ should make the values of $R_L\left(u_{\theta}\left(x,t\right)\right)$, $R_B\left(u_{\theta}\left(x,t\right)\right)$, and $R_I\left(u_{\theta}\left(x,t\right)\right)$ equal to $0$.

For training $u_{\theta}$, the first step is to generate training points. Since \eqref{eqn}, \eqref{boundary}, and \eqref{initial} respectively describe the governing equations, boundary conditions, and initial conditions of the initial-boundary value problem, the selection of training points must be conducted separately.
Specifically, for the governing equations \eqref{eqn}, training points need to be chosen within $\Omega \times (0, T]$, denoted as $\left\{x_L^j, t_L^j\right\}_{j=1}^{N_L}$.
Similarly, for the boundary conditions \eqref{boundary} and the initial conditions \eqref{initial}, training points are selected on $\partial\Omega \times (0, T]$ and $\bar{\Omega} \times \{0\}$ respectively, denoted as $\left\{x_B^j, t_B^j\right\}_{j=1}^{N_B}$ and $\left\{x_I^j, t_I^j\right\}_{j=1}^{N_I}$. Here, $N_L$, $N_B$, and $N_I$ represent the respective numbers of training points chosen for the governing equations, boundary conditions, and initial conditions.

In the training process of PINNs, we minimize the following loss function
\begin{flalign*}
    &\ \ \ \ \ \ \ \ loss:={loss}_L + {\lambda}_B {loss}_B + {\lambda}_I {loss}_I,&
\end{flalign*}
where ${\lambda}_B$ and ${\lambda}_I$ represent the loss weights for boundary conditions and initial conditions, respectively,
\begin{flalign*}
    &\ \ \ \ \ \ \ \ loss_L:=\frac{1}{N_L}\sum_{j=1}^{N_L}\left(R_L\left(u_{\theta}\left(x_L^j,t_L^j\right)\right)\right)^2,\
     loss_B:=\frac{1}{N_B}\sum_{j=1}^{N_B}\left(R_B\left(u_{\theta}\left(x_B^j,t_B^j\right)\right)\right)^2,\ &  
\end{flalign*}
and
\begin{flalign*}
    &\ \ \ \ \ \ \ \ loss_I:=\frac{1}{N_I}\sum_{j=1}^{N_I}\left(R_I\left(u_{\theta}\left(x_I^j,t_I^j\right)\right)\right)^2.&
\end{flalign*}

\begin{figure}
    \centering
    \includegraphics[width=0.8\textwidth]{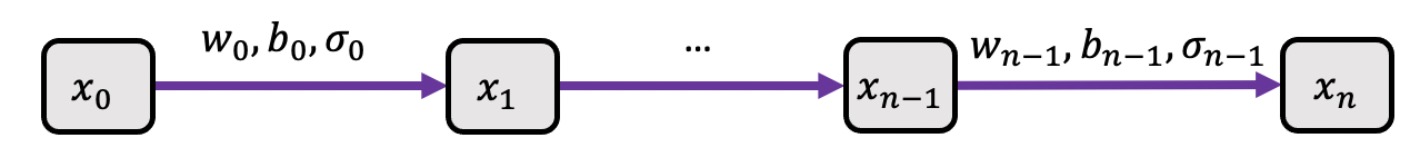}
    \caption{The architecture of a FNN. The purple arrows depict the fully connected relationships between different components.}
    \label{fig:FC}
\end{figure}

The neural network architecture employed in PINNs consists of fully connected layers, illustrated in Fig.\ \ref{fig:FC}. In this representation, $w_i$, $b_i$, and $\sigma_i$ correspond to the weight, bias, and activation function of the $i$-th hidden layer $(i=0,1,\ldots,n-2)$. Here, $x_0$ represents the input to the neural network, $x_n$ is the output, and $x_i$ signifies the output of the $i$-th hidden layer ($i=0,1,\ldots,n-2$). It's important to note that $x_i$ ($i=0,1,\ldots,n$) is not necessarily a scalar, but can also denote a high-dimensional vector. The forward propagation process of the FNN can be expressed as follows:
\begin{flalign}
    \label{fc_math}
    &\ \ \ \ \ \ \ \ x_i=\sigma_{i-1}\ (w_{i-1} x_{i-1} + b_{i-1}\ ),\ \ i=1,2,\ldots,n.&
\end{flalign}

When the solutions of partial differential equations do not exhibit drastic changes, indicating gradual variations, PINNs constructed with regular FNNs are effective in solving the problems.
However, challenges arise when dealing with PDEs characterized by rapidly-changing solutions.
In such cases, PINNs face issues, including slow initial convergence of the loss function during training, abrupt spikes in the loss function that are hard to reduce, and a susceptibility to becoming trapped in local minima, ultimately resulting in reduced solution accuracy.
We propose BsPINNs, which are introduced in detail in the following section, to address these challenges encountered by PINNs.

\section{Binary Structured Physics-Informed Neural Networks}\label{s3}
In this section, we will introduce binary structured physical information embedding neural networks, which can be used to solve the PDEs characterized by rapidly-changing solutions. To do so, we will first clarify the motivation and key ideas behind addressing such challenging problems. Subsequently, we will explore an innovative network structure - binary structured neural networks (BsNNs), which is an internal neural network component of BsPINNs.
Finally, we will illustrate the characteristics of BsNNs through a specific example. It is important to note that, apart from the difference in neural network structure compared to PINNs, all other aspects of BsPINNs remain unchanged, including the construction of the loss function, parameter update methods, and more.

\subsection{Motivations and inspirations}\label{s_motivation}
A substantial body of existing literature indicates that PINNs are not suitable for solving PDEs characterized by rapidly-changing solutions \cite{coutinho2023physics, jagtap2020conservative, wight2020solving, zeng2022adaptive}.

Indeed, the network parameters of FNNs are globally correlated with the training data used during the network's training. This correlation implies that any modification in a set of training data will induce changes in all network parameters. Consequently, when employing an FNN to approximate a target function, adjustments in the network parameters that enhance the accuracy of approximation in a localized region may simultaneously increase the overall error in approximating the entire target function. This behavior can result in the training process of an FNN getting stuck in a local minimum while optimizing the loss function. This phenomenon becomes particularly evident when the target function exhibits rapid changes.

\begin{figure}[h!]
    \centering
    \begin{subfigure}{0.4\textwidth}
        \includegraphics[width=\textwidth]{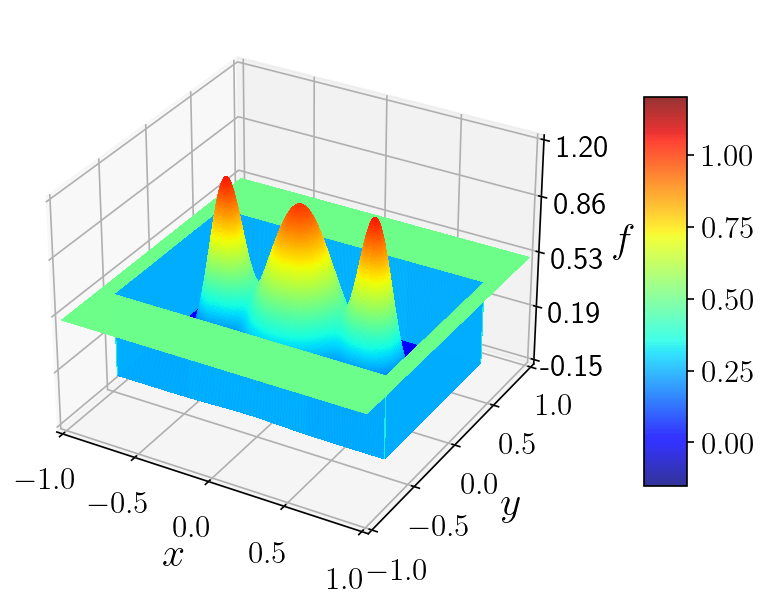}
        \caption{}
    \end{subfigure}
    \begin{subfigure}{0.4\textwidth}
        \includegraphics[width=\textwidth]{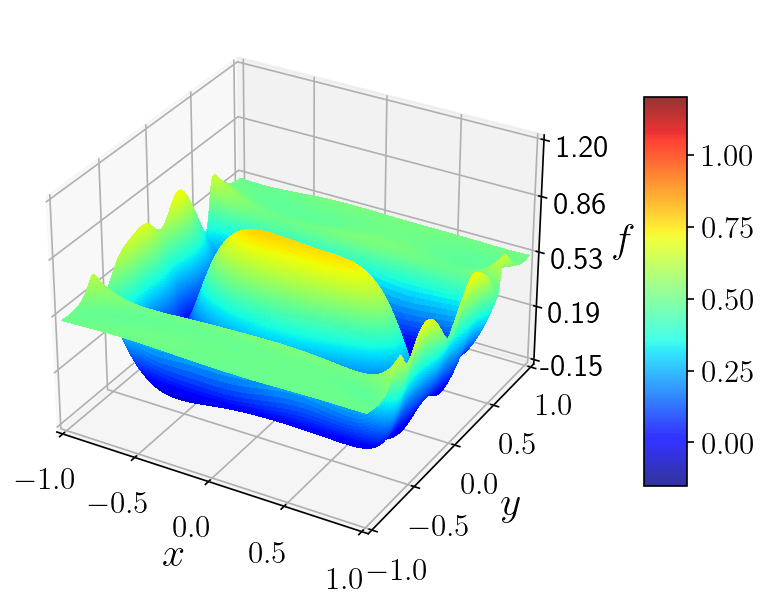}
        \caption{}
    \end{subfigure}
    \caption{Function $f$ on $\Omega$ (a) and the FNN-based approximation of function $f$ (b).}
        \label{fig:funv2_fc}
\end{figure}

For instance, let's consider the use of a FNN with four hidden layers, each comprising $32$ neurons, to approximate the following function:
\begin{flalign}\label{eq_test_f}
    &\ \ \ \ \ \ \ \ f(x,y)=
    \begin{cases}
        e^{-\left(\frac{(x+0.5)^2}{0.02}+\frac{y^2}{0.02}\right)} + e^{-\left(\frac{x^2}{0.08}+\frac{y^2}{0.08}\right)} + e^{-\left(\frac{(x-0.5)^2}{0.02}+\frac{y^2}{0.02}\right)}, & (x,y) \in \Omega_1, \\
        0.5, & (x,y) \in \Omega \backslash \Omega_1,
    \end{cases}&
\end{flalign}
where
\begin{flalign*}
    &\ \ \ \ \ \ \ \ \Omega = [-1,1]^2, &\\
    &\ \ \ \ \ \ \ \ \Omega_1 = [-0.9,0.9] \times [-0.6,0.6]. &
\end{flalign*}
We employ $15,000$ randomly sampled training points within $\Omega$, utilize the sigmoid activation function, and terminate the training after $10,000$ iterations. As depicted in image (b) of Fig.\ \ref{fig:funv2_fc}, the FNN-based approximation of function $f$ captures the general trend of the peaks but fails to represent the three distinct peaks individually. Instead, it forms a single, larger peak. Similarly, in areas where the function $f$ exhibits discontinuities, the FNN-based approximation portrays the function $f$ smoothly, but it fails to capture the jumps at the discontinuity points. This example illustrates that FNNs tend to perform poorly in accurately learning local features located in rapidly changing regions of functions.

This observation naturally leads us to draw an analogy with approximating functions using Fourier series, a global approximation method renowned for capturing the overall characteristics of functions but limited in its ability to represent local features \cite{trigub2004fourier}. As demonstrated by the Gibbs phenomenon \cite{gottlieb1997gibbs}, the Fourier series of a function with a jump discontinuity tends to perform poorly. To better capture the local characteristics of functions, many researchers turn to wavelet series \cite{daubechies1992ten} which excel at approximating non-stationary functions or functions with rapidly changing local behaviors \cite{mallat1989theory}. This is because the coefficients of wavelet series are only related to the local values of functions. When approximating a function using wavelet series, a change in the function's value at a specific point affects only a subset of the coefficients of wavelet series, rather than all of them.

This insight guides us in designing neural network structures suitable for capturing local features: neurons in the subsequent layer should be connected only to a subset of neurons in the previous layer.

The aim to design a network for learning the local features of functions also brings to mind the idea of a ``mixture of experts'' (MoEs) model \cite{jacobs1991adaptive}, where the model comprises multiple independent expert networks. Each expert specializes in solving a subproblem within a complex task, and their collective knowledge is combined to address the overall complex problem.
In this context, we can design a neural network model similar to MoE, composed of multiple sub-networks, with each sub-network dedicated to learning a specific local feature of the solution. The collective knowledge gained by these sub-networks represents the complete set of solution features. Simultaneously, to facilitate coordination among the sub-networks and ensure the sensible allocation of the task of learning local features, we arrange for the sub-networks to share some parameters in the initial hidden layers.

By amalgamating these concepts, we establish the BsNNs within BsPINNs. In the next subsection, we provide a detailed presentation of this architecture.

\subsection{Binary structured network architecture} \label{s3.2}
\begin{figure}[h!]
    \centering
    \begin{subfigure}{0.59\textwidth}
        \includegraphics[width=\textwidth]{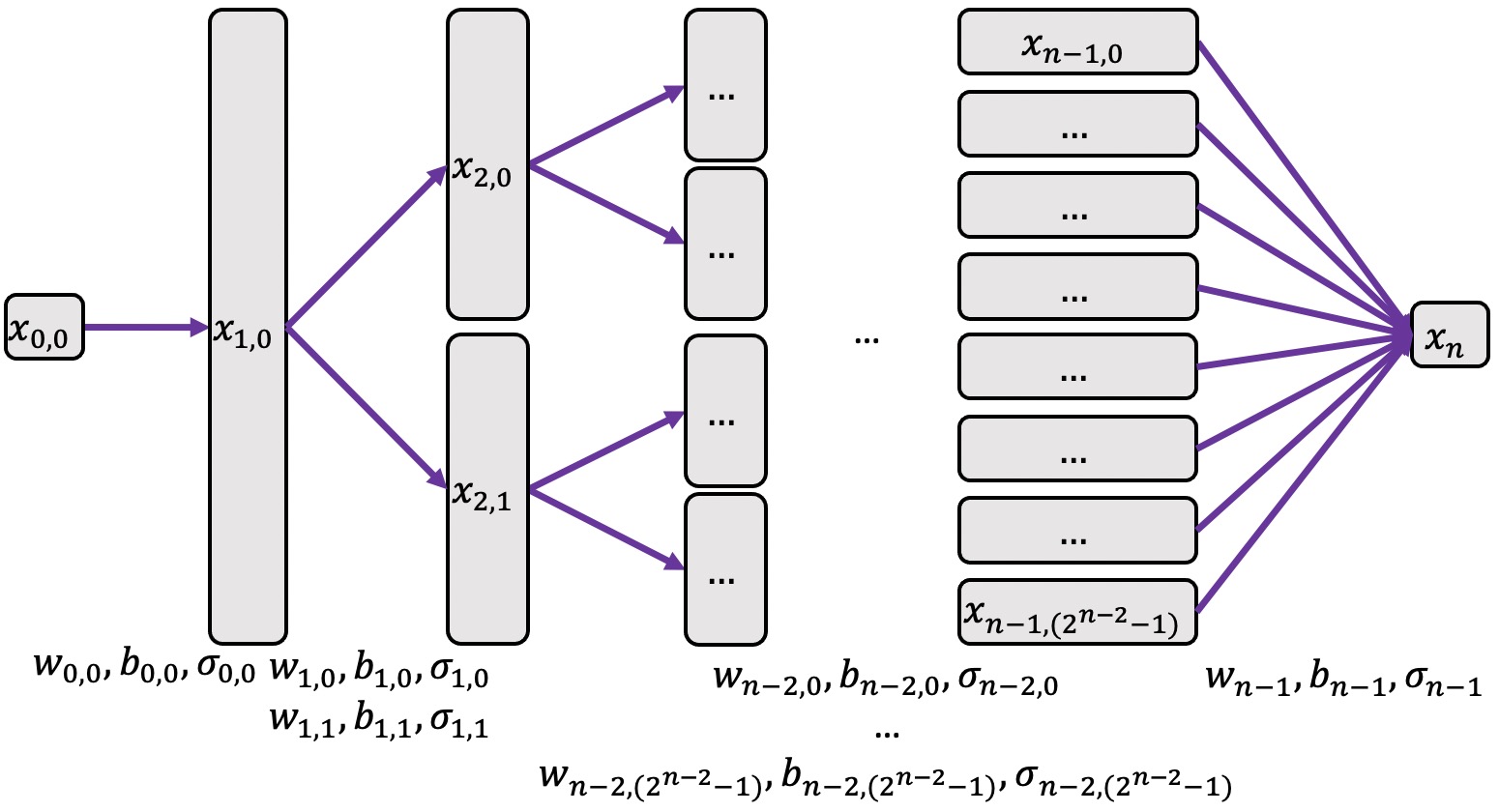}
        \caption{}
    \end{subfigure}
    \begin{subfigure}{0.39\textwidth}
        \includegraphics[width=\textwidth]{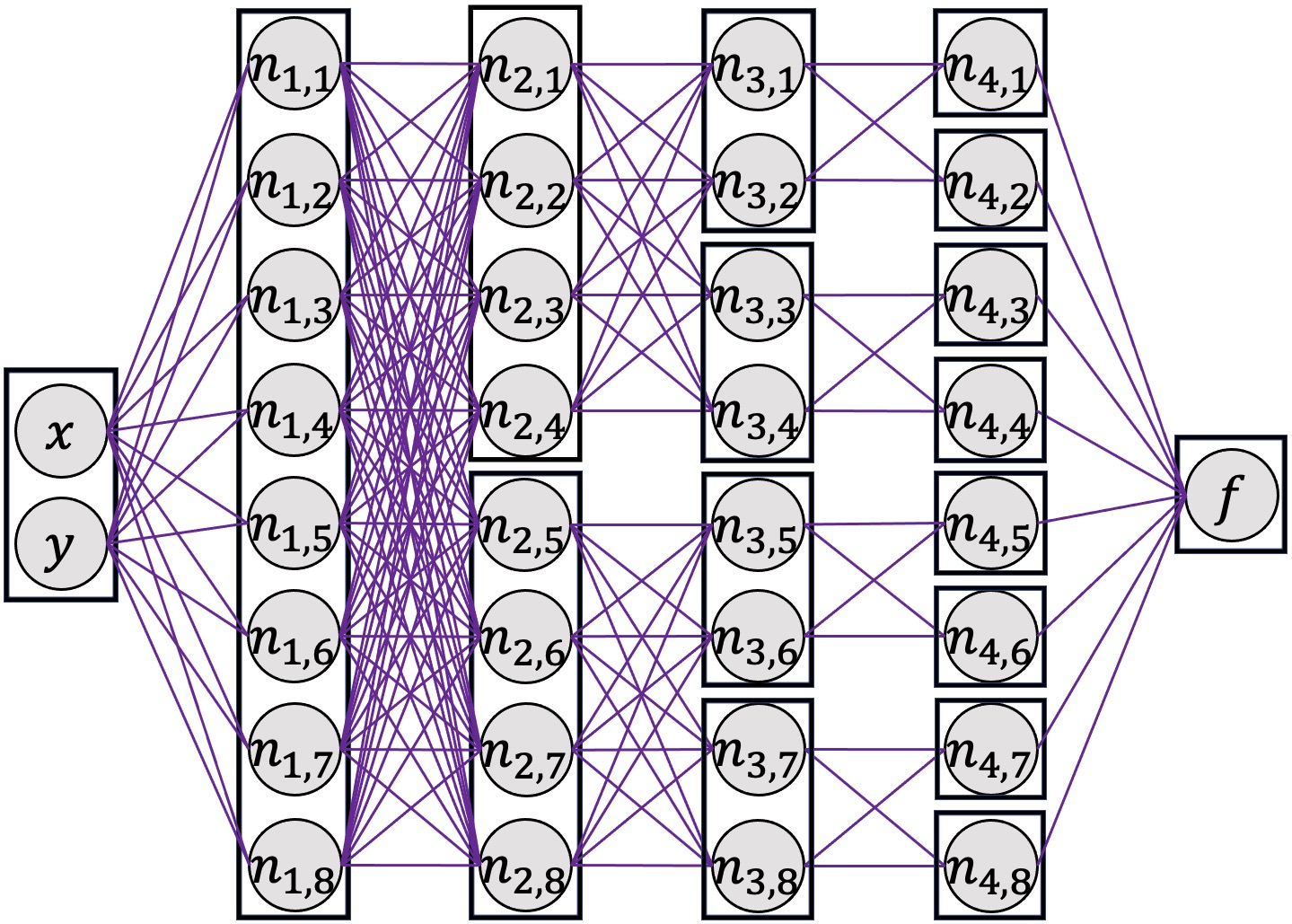}
        \caption{}
    \end{subfigure}
    \caption{(a) The general network structure of a BsNN. (b) The structure of a BsNN consists of $4$ hidden layers, each with $8$ neurons.}
    \label{fig:BFC}
\end{figure}

Building upon the FNN architecture in equation \eqref{fc_math}, we introduce a novel network structure known as the Binary Structured Neural Network (BsNN), illustrated in Fig.\ \ref{fig:BFC}(a).
Each $x_{i,j}$ in Fig.\ \ref{fig:BFC}(a) contains one or more neurons, and such $x_{i,j}$ is referred to as a ``neuron block''.
The purple arrows indicate fully connected relationship between two neuron blocks, reflecting trainable weight parameters connecting every neuron pair from the two neural blocks.
Within the final layer, the outputs of each neuron block are concatenated and fully connected to the output $x_n$.
This structure resembles a binary tree, where the neuron blocks in each hidden layer, except the first and the last, are fully connected to two neuron blocks in the next hidden layer.

In this network, $w_{i,j}$, $b_{i,j}$, and $\sigma_{i,j}$ represent the weight, bias, and activation function, respectively, for the $j$-th branch of the $i$-th hidden layer ($j=0,1,\ldots,2^i-1$, $i=0,1,\ldots,n-2$). Correspondingly, $w_{n-1}$, $b_{n-1}$, and $\sigma_{n-1}$ denote the weight, bias, and activation function for the final layer.
Each neuron block within the same hidden layer contains the same number of neurons, and this quantity is known as the ``block size'' of that hidden layer. In this paper, the number of neuron blocks for the $i$-th layer with $1\leq i<n$ is $2^{i-1}$, and for $1<i<n$, the block size in the $(i-1)$-th layer is twice that of the $i$-th layer.  Throughout this paper, we will consistently use the terms ``neuron block" and ``block size".

With these notations, the forward propagation of the BsNN can be mathematically expressed as follows:
\begin{flalign*}
    &\ \ \ \ \ \ \ \ x_{i,j}=\sigma_{i-1,j}\left(w_{i-1,j} x_{i-1,\left\lfloor\frac{j}{2}\right\rfloor} + b_{i-1,j}\right),\ i=1,\ldots,(n-1),\ j=0,\ldots,\left(2^{i-2}-1\right). &
\end{flalign*}
The outputs $x_{n-1,j}$ (where $j=0,1,\ldots,(2^{n-2}-1)$) are concatenated along the last dimension to create a variable referred to as $x_{n-1}$. Subsequently, the final output of the BsNN is obtained as follows
\begin{flalign*}
    &\ \ \ \ \ \ \ \ x_n=\sigma_{n-1}(w_{n-1} x_{n-1} + b_{n-1}).&
\end{flalign*}
Notably, when the network possesses substantial depth, the parameter count of a BsNN is notably lower than that of an equivalently sized FNN featuring the same quantity of neurons.

To provide a more intuitive explanation of this architecture, let's consider a BsNN comprising four hidden layers, each containing eight neurons, as depicted in Fig.\ \ref{fig:BFC} (b).
In this diagram, the variables $x$ and $y$ represent the network inputs, while $f$ represents the network output.
The block size for each hidden layer is $8$, $4$, $2$, and $1$, respectively.
The black squares symbolize the previously defined neuron blocks.
The corresponding FNN with four hidden layers, each containing eight neurons, has $3,297$ trainable parameters.
In contrast, the BsNN in Fig.\ \ref{fig:BFC}(b) has $2,017$ trainable parameters.
Therefore, the BsNN has only $61.1\%$ of the parameters of the corresponding FNN, which significantly improves training speed and conserves memory.
In subsequent discussions, the BsNN shown in Fig.\ \ref{fig:BFC} (b) will be denoted as ``8-1'', and the corresponding FNN will be denoted as ``4*8''.
We will use this abbreviation notation throughout this paper.

To validate the performance of the BsNN, we approximate the function $f$ defined in \eqref{eq_test_f} using the BsNN. This BsNN has the same number of neurons and hidden layers as the FFN in subsection \ref{s_motivation}, which consists of $4$ hidden layers with block sizes of $32$, $16$, $8$, and $4$, respectively. All other hyperparameters remain consistent with those in subsection \ref{s_motivation}.

\begin{figure}[h!]
    \centering
    \includegraphics[width=.5\textwidth]{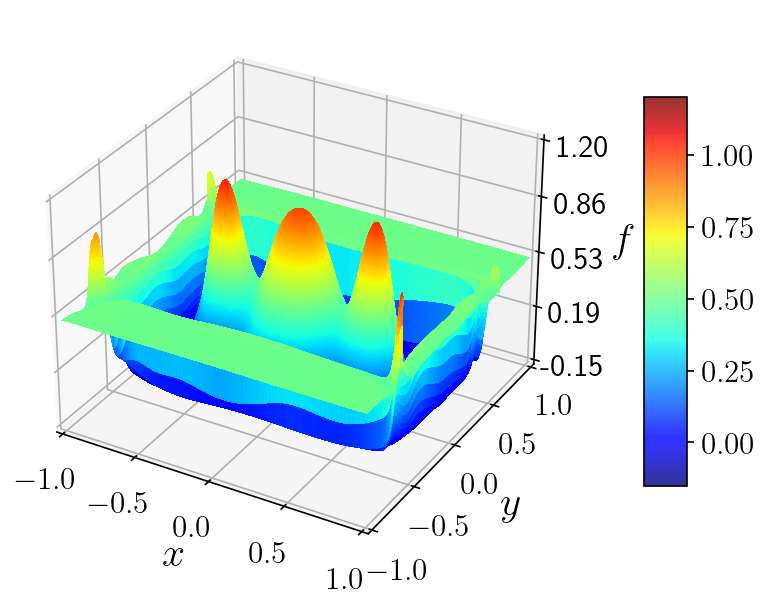}
    \caption{The BsNN-based approximation of function $f$.}
    \label{fig:funv2_bfc}
\end{figure}

Comparing Fig.\ \ref{fig:funv2_fc}(b) and Fig.\ \ref{fig:funv2_bfc}, we can observe that the BsNN effectively captures the presence of three peaks, as shown in Fig.\ \ref{fig:funv2_fc}(a), and provides a good approximation for numerical discontinuities. This performance is significantly better than that of the FNN.

\begin{figure}[h!]
    \centering
    \includegraphics[width=0.4\textwidth]{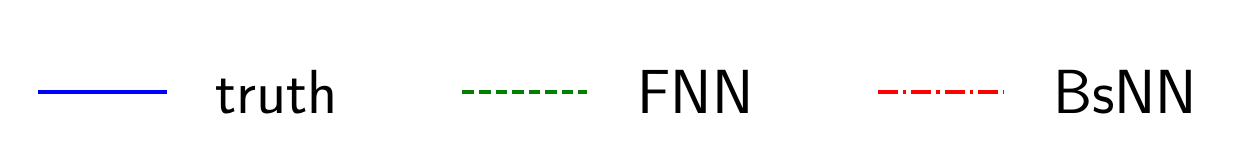} \\
    \begin{subfigure}{0.4\textwidth}
        \includegraphics[width=\textwidth]{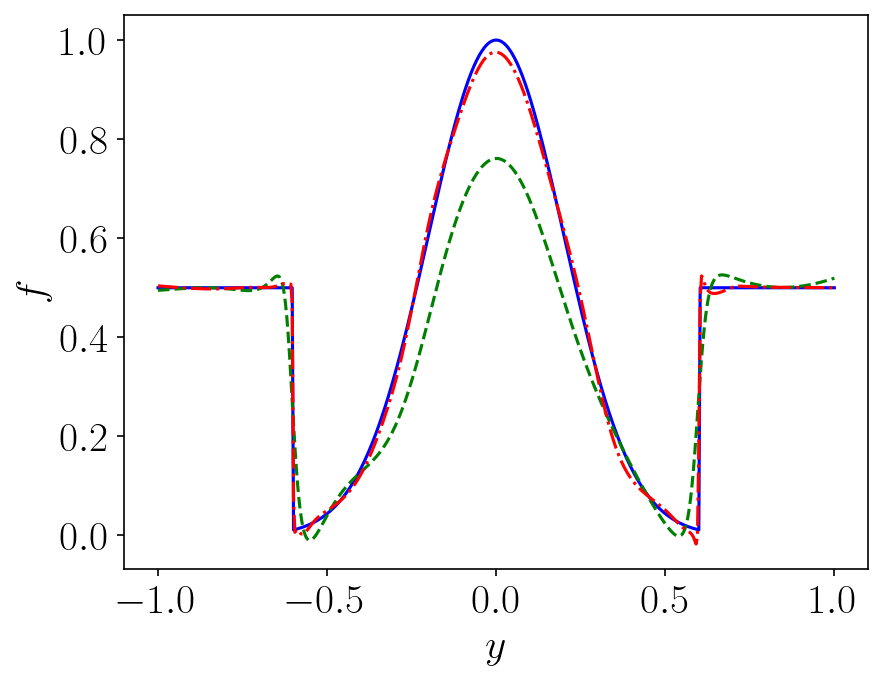}
        \caption{Cross-sections by the plane $x = 0$}
    \end{subfigure}
    \begin{subfigure}{0.4\textwidth}
        \includegraphics[width=\textwidth]{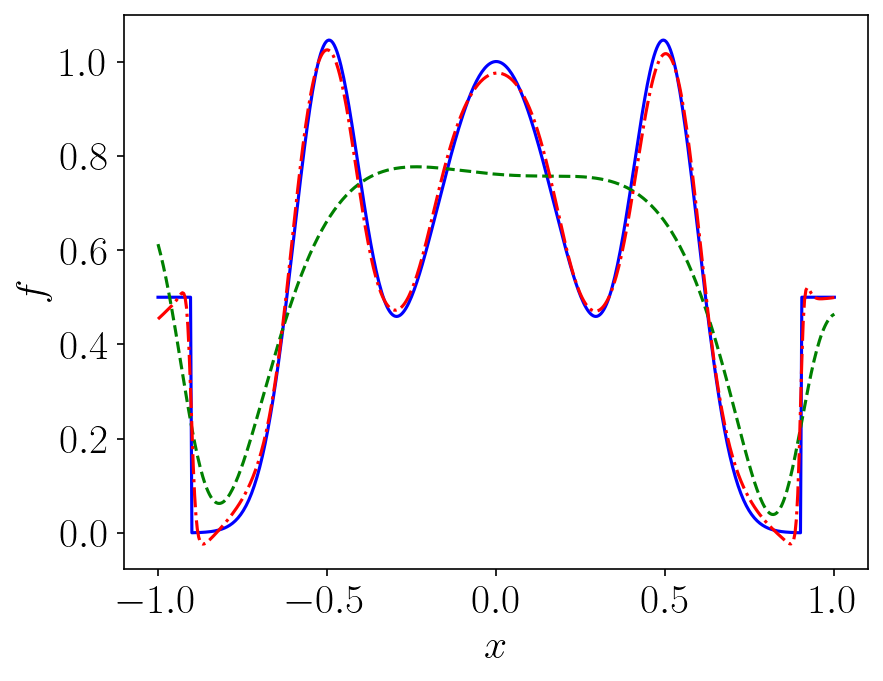}
        \caption{Cross-sections by the plane $y=0$}
    \end{subfigure}
    \caption{Cross-sections of the function $f$ and its approximations by the FNN and BsNN on the planes $x=0$ (a) and $y=0$ (b). The blue solid lines represent the true values of $f$ , the green dashed lines represent the predicted values by the FNN, and the green dot-dashed lines represent the predicted values by the BsNN.}
    \label{fig:funv2_cross}
\end{figure}

To facilitate a clearer comparison of the approximation performance of the two neural networks, we generate cross-sectional plots at $x=0$ and $y=0$, as shown in Fig.\ \ref{fig:funv2_cross}. In Fig.\ \ref{fig:funv2_cross}, the blue solid lines represent the true values of $f$, the green dashed lines represent the predicted values by the FNN, and the green dot-dashed lines represent the predicted values by the BsNN. As depicted in Fig.\ \ref{fig:funv2_cross}(a) and (b), the FNN fails to precisely capture the discontinuities. The BsNN exhibits more significant variations in areas of function discontinuities and can more accurately represent the function's discontinuities. Additionally, in Fig.\ \ref{fig:funv2_cross}(b), we also observe that the FNN approximates the three central peaks in a relatively smooth manner and does not capture the oscillatory nature between the peaks. In contrast, the BsNN performs much better in approximating the three peaks.

\begin{figure}[h!]
    \centering
    \includegraphics[width=.4\textwidth]{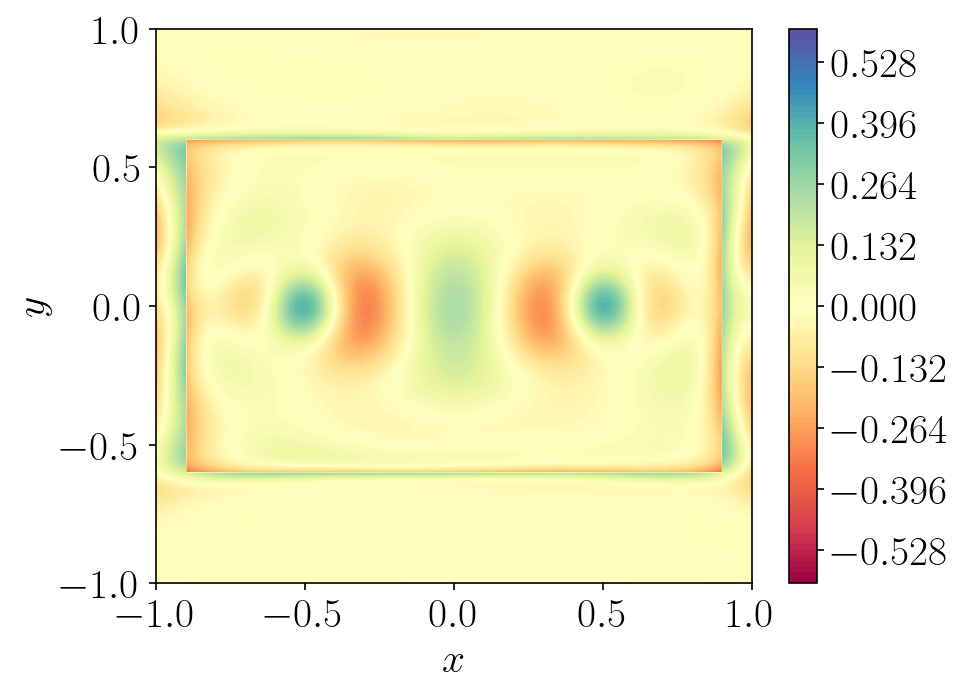}
    \includegraphics[width=.4\textwidth]{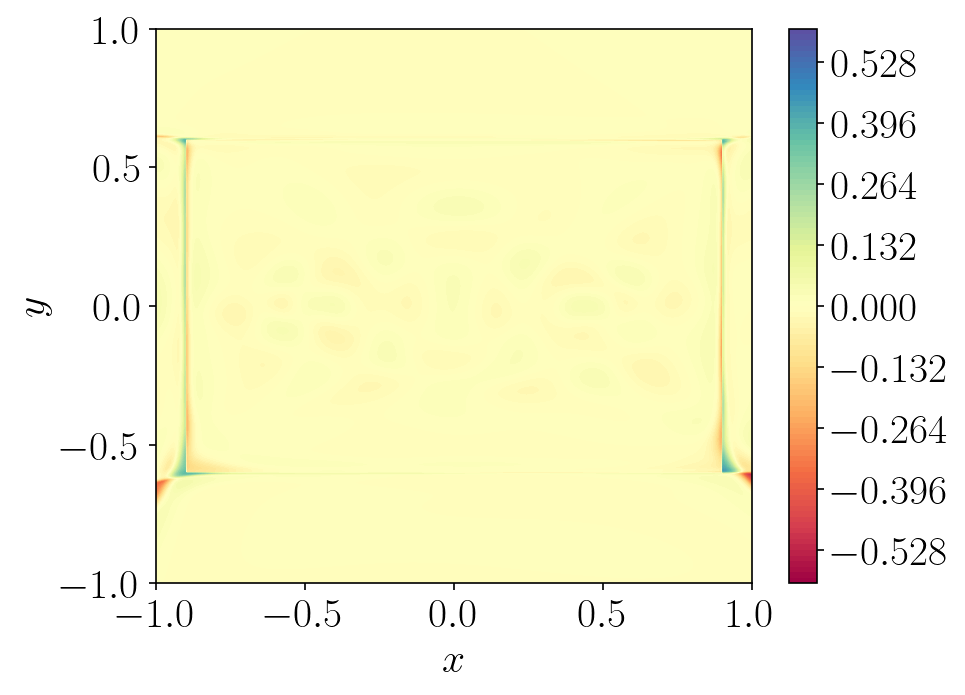}
    \caption{Point-wise errors in function approximation for the FNN (left) and the BsNN (right).}
    \label{fig:funv2_error}
\end{figure}

The advantages of the BsNN become even more apparent when examining the error comparison in Fig.\ \ref{fig:funv2_error}, where the left image presents the differences between $f$ and the predicted values of the FNN, and the right image illustrates the differences between $f$ and the predicted values of the BsNN. It's evident that the differences between $f$ and the predicted values of the BsNN are much smaller than the differences between $f$ and the predicted values of the FNN. In particular, near the discontinuity points, the error of the BsNN is significantly smaller than that of the FNN.

In the following sections, we will explore the performance of BsPINNs with BsNNs in solving various PDEs.

\section{Numerical Examples} \label{s4}
In this section, we evaluate the performance of BsPINNs on various PDEs, and also employ PINNs to solve the same set of PDEs for comparison.
In subsection \ref{burgers_one}, we solve the one-dimensional Burgers equation with a continuous solution that exhibits rapid changes.
In subsection \ref{euler_two}, we solve the Euler equation with discontinuous solution and analyze the results from the perspective of long-tailed distribution.
In subsection \ref{s_highhe}, we resolve the two-dimensional Helmholtz equation and analyze the outputs of each channel in a BsPINN to investigate how BsPINNs utilize mechanisms similar to MoEs to achieve improved solving performance.
In subsection \ref{he3d}, we solve the three-dimensional Helmholtz equation in a 3D M\"{o}bius knot and explore the impact of varying numbers of training points on model performance.
In subsection \ref{s_pos}, we resolve the high-dimensional oscillatory Poisson equation and delve into the impact of various network architectures on the model performance.

Our deep learning framework of choice is PyTorch version 1.10.0+cu113, and we adopt the Adam optimizer \cite{Kingma_Ba_2014} to optimize the network parameters.
For learning rate scheduler in subsections \ref{burgers_one} to \ref{he3d}, we use available in Pytorch ReduceLROnPlateau with the patience of one-tenth of the training iterations and learning rate decay factor of $0.5$. In subsection \ref{s_pos}, we use the exponential decay learning rate strategy, that is, every $1000$ epochs, the learning rate becomes 0.95 times the original, which is consistent with \cite{zeng2022adaptive}.
For all PDEs, the training points randomly selected follow uniform distribution within their corresponding domains.
Our computational resources are powered by an NVIDIA GeForce RTX 3090 GPU.

To ensure the robustness of our method and reduce the impact of random variation, we perform numerical experiments without loss of generality by repeating them 10 times with different random seeds for selecting training points and initializing the neural network parameters.
During each training run, the model corresponding to the iteration step with the minimum loss value is selected as the final output model of the training process, which will be utilized for the computation of relative error.
In subsections \ref{burgers_one} to \ref{s_highhe}, when calculating the relative error, we discretize the solution domain into a uniform grid and then compute the relative error at the grid nodes.
Here, dividing a two-dimensional region into an $m \times n$ uniform grid entails uniform division along the first dimension using $m$ nodes and the second dimension using $n$ nodes. Consequently, there will be a total of $m \times n$ grid nodes, rather than $m \times n$ units. This extends similarly to three-dimensional uniform grids.
In this section, we use the format ``\emph{average $\pm$ standard deviation}" to represent the average and standard deviation of the relative errors from $10$ runs for each experiment.
In each experiment, the images of the predicted solutions presented are obtained from the model with the minimum relative error, which is selected from the $10$ runs with different seeds, providing a representative illustration of the model’s performance.

\subsection{One-dimensional Burgers equation} \label{burgers_one}
In this section, we consider the one-dimensional Burgers equation, which is expressed as
\begin{flalign*}
    &\ \ \ \ \ \ \ \ \frac{\partial u}{\partial t} + u\frac{\partial u}{\partial x} - \left( \frac{0.01}{\pi} \right) \frac{\partial^2 u}{\partial x^2} = 0, x \in (-1,1), t \in (0,1],&\\
    &\ \ \ \ \ \ \ \ u(x,0) = -\sin(\pi x), x \in [-1,1],&\\
    &\ \ \ \ \ \ \ \ u(-1,t) = u(1,t) = 0, t \in (0,1],&
\end{flalign*}
where $u$ represents the displacement.

We conduct a comparative analysis between two neural network architectures: ``5*256'' and ``256-16'' corresponding to PINNs and BsPINNs respectively. Within the neural network architecture, the tanh activation function is utilized
in all hidden layers, except for the ﬁnal layer, where a linear activation function is employed.
For training points corresponding to the governing equation, we sample $30,000$ training points using latin hypercube design over the domain $(-1, 1) \times (0, 1]$.
The adoption of the latin hypercube design sampling method here follows the same approach as in \cite{mcclenny2020self}. For training points corresponding to the initial condition, we randomly sample $200$ points across $[-1,1] \times \{0\}$. For training points corresponding to the boundary condition, we randomly sample $100$ points at each of the two boundaries, $x = -1$ and $x = 1$.
Following the notations introduced in Section \ref{s2}, we set ${\lambda}_I={\lambda}_B=1$ in the loss function, which remains consistent with the configuration in \cite{mcclenny2020self}.
We fix the number of iterations to $10,000$ and set an initial learning rate of $0.005$.

\begin{figure}[h!]
    \centering
    \includegraphics[width=.4\textwidth]{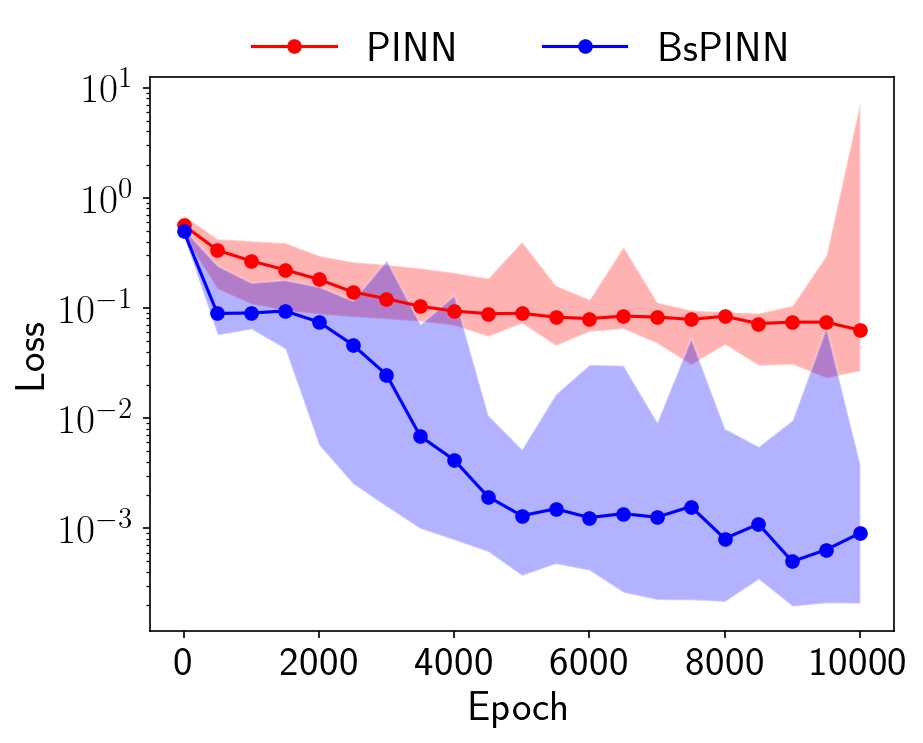}
    \caption{The dynamic change curves of the training loss for the PINNs and BsPINNs when solving the one-dimensional Burgers equation. The shaded regions here represent the segment between the maximum and minimum values of the loss at specific iteration steps from $10$ runs with different seeds, while the circular nodes represent the median values.}
    \label{fig:burgers_loss}
\end{figure}

Depicted in Fig.\ \ref{fig:burgers_loss} is the dynamic change of the loss values with respect to the training steps. After around $3,000$ epochs, the median loss-value curve of the PINNs exhibits a conspicuous plateau, stabilizing at approximately $0.1$. Consequently, achieving substantial decreases in the loss values becomes challenging. In contrast, the median loss values of the BsPINNs briefly stall around $0.1$ in the early stages of training before rapidly decreasing thereafter. This significant difference shown in Fig.\ \ref{fig:burgers_loss} reflects the considerable advantage of BsPINNs in convergence speed compared to PINNs.

During the error assessment, the spatio-temporal domain $[-1,1] \times [0,1]$ is partitioned into a uniform grid of $256 \times 100$, which divides the spatial domain $[-1, 1]$ into a uniform grid using $256$ nodes and the temporal domain $[0,1]$ into a uniform grid using $100$ nodes.
By extracting the grid nodes from this $256 \times 100$ grid, we obtain 25,600 points, denoted as $(x_i,t_i)(i=1,2,\ldots,N)$ where $N=25,600$.
Subsequently, the relative error is calculated as
\begin{flalign*}
    &\ \ \ \ \ \ \ \ \text{error} = \frac{\sqrt{\sum_{i=1}^{N}\left(u_{\theta}\left(x_i,t_i\right)-u^\ast\left(x_i,t_i\right)\right)^2}}{\sqrt{\sum_{i=1}^{N}\left(u^\ast\left(x_i,t_i\right)\right)^2}},&
\end{flalign*}
where $u_{\theta}$ represents the predicted solution computed by neural network parameterized by $\theta$, and $u^\ast$ denotes the reference solution.
The relative error of the PINNs is $2.686 \times 10^{-1} \pm 6.285 \times 10^{-2}$, whereas the BsPINNs exhibit a relative error of $1.252 \times 10^{-2} \pm 2.926 \times 10^{-3}$.
This indicates that the average relative error of the BsPINNs is less than one-tenth of that of the PINNs, and the standard deviation of the relative errors for the BsPINNs is also less than one-tenth of that for the PINNs. These results demonstrate the higher precision and more stable performance of BsPINNs in solving the one-dimensional Burgers equation.

\begin{figure}[h!]
    \centering
    \includegraphics[width=0.3\textwidth]{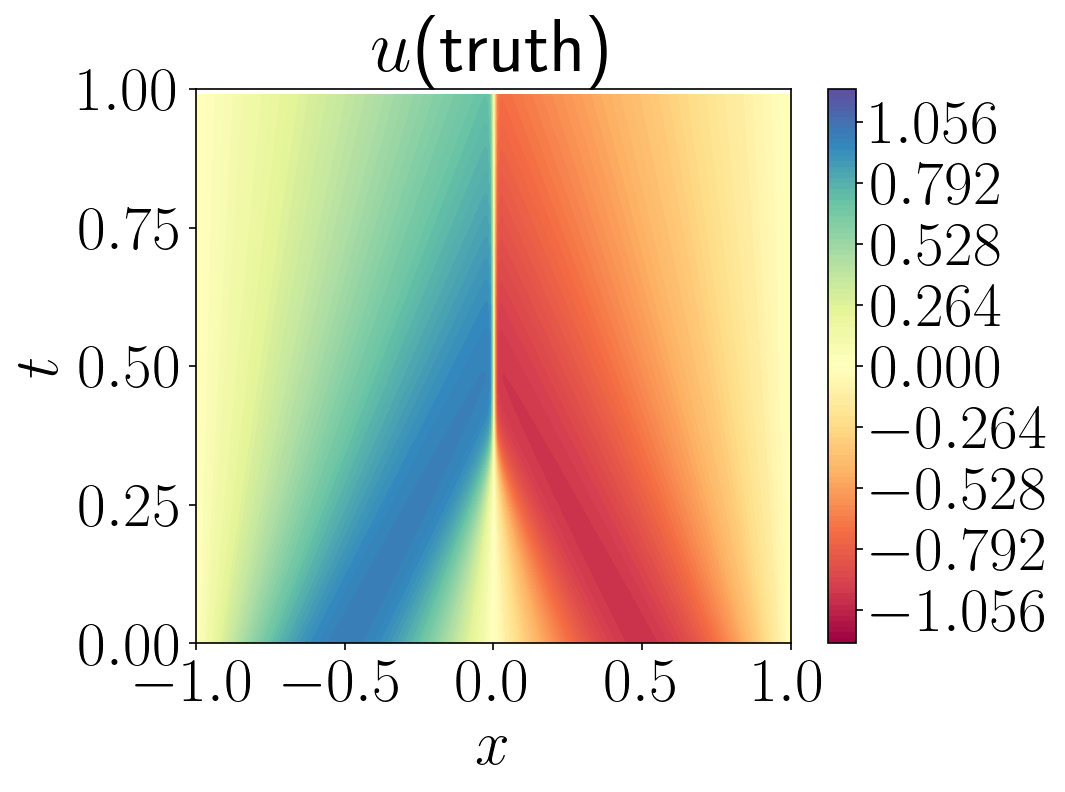}
    \includegraphics[width=0.3\textwidth]{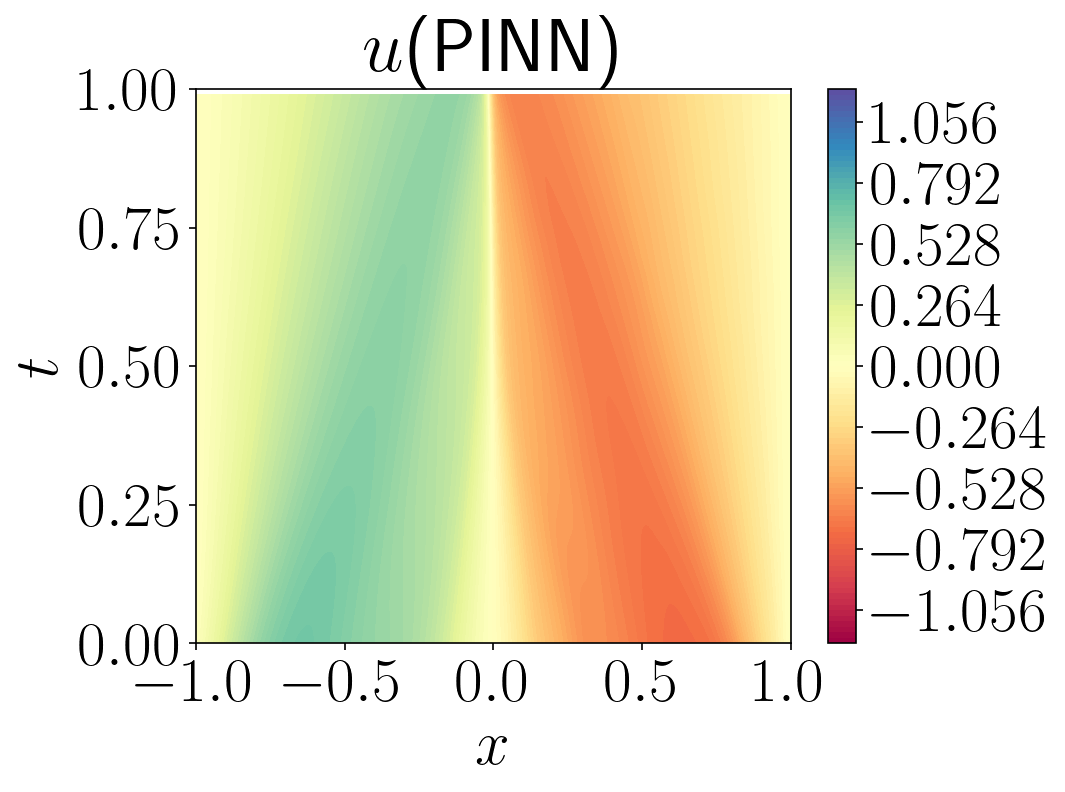}
    \includegraphics[width=0.3\textwidth]{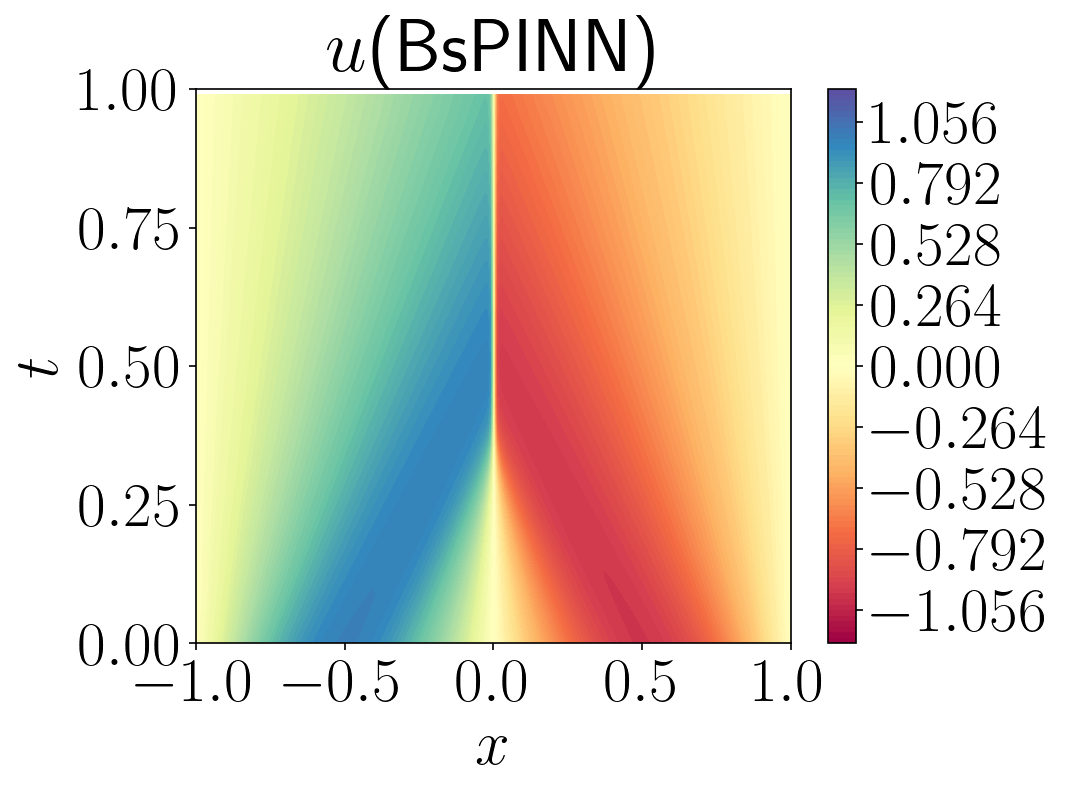}
    \caption{The reference solution (left) and the predicted solutions of the PINN (middle) and the BsPINN (right) for the one-dimensional Burgers equation.}
    \label{fig:burgers_u}
\end{figure}

The contrast between PINNs and BsPINNs is highlighted in Fig.\ \ref{fig:burgers_u}, where we present the reference solution alongside the predicted solutions obtained from the PINN and BsPINN, each derived from the respective run with the minimum relative error out of $10$.
The PINN fails to capture the rapid changes in values near $x=0$, opting for a smoother fit instead of accurately modeling the sharp variations. Additionally, the absolute values of the maximum and minimum values in the predicted solution of the PINN are much smaller than those in the reference solution, reflecting the characteristic that the PINN’s predicted solution tends to be smooth.
In contrast, the BsPINN effectively captures the rapid changes in values near $x=0$, and the absolute values of the maximum and minimum predicted values are very close to the reference solution, achieving a higher level of accuracy.

Through the experiment solving the one-dimensional Burgers equation, it is evident that BsPINNs hold a significant advantage over PINNs in terms of convergence speed and solution accuracy.

\subsection{Two-dimensional Euler equation} \label{euler_two}
In this subsection, we delve into the Euler equation governing two-dimensional compressible flow, which can be expressed as
\begin{flalign*}\label{eq_euler}
    &\ \ \ \ \ \ \ \ \frac{\partial\rho}{\partial t}+\frac{\partial(\rho u)}{\partial x}+\frac{\partial(\rho v)}{\partial y}=0, &\\
    &\ \ \ \ \ \ \ \ \frac{\partial\left(\rho u\right)}{\partial t}+\frac{\partial\left(\rho u^2+p\right)}{\partial x}+\frac{\partial\left(\rho uv\right)}{\partial y}=0, &\\
    &\ \ \ \ \ \ \ \ \frac{\partial\left(\rho v\right)}{\partial t}+\frac{\partial\left(\rho uv\right)}{\partial x}+\frac{\partial\left(\rho v^2+p\right)}{\partial y}=0, &\\
    &\ \ \ \ \ \ \ \ \frac{\partial\left(\rho E\right)}{\partial t}+\frac{\partial\left(u\left(\rho E+p\right)\right)}{\partial x}+\frac{\partial\left(v\left(\rho E+p\right)\right)}{\partial y}=0, &\\
    &\ \ \ \ \ \ \ \ p=\left(\gamma-1\right)\left(\rho E-\frac{1}{2}\rho\left(u^2+v^2\right)\right),&
\end{flalign*}
where $(x,y)\in(0,1)^2$ represents the spatial variable, $t\in(0,2]$ represents the temporal variable, $\rho$ represents the fluid density, $u$ represents the fluid velocity in $x$-direction, $v$ represents the fluid velocity in $y$-direction, and $p$ represents the pressure.

An initial shock is given at $x = 0.5, 0 \le y \le 1$, and the initial conditions would be
\begin{flalign*}
    &\ \ \ \ \ \ \ \ \rho(x,y,0)=
    \begin{cases}
        1.4, & 0\le x\le0.5,\ 0\le y\le1, \\
        1.0, & 0.5<x\le1,\ 0\le y\le1,
    \end{cases} &\\
    &\ \ \ \ \ \ \ \ u(x,y,0)=0.1,\ v(x,y,0)=0,\ p(x,y,0)=1, (x,y)\in[0,1]^2.&
\end{flalign*}
Moreover, we apply the following Dirichlet boundary conditions
\begin{flalign*}
    &\ \ \ \ \ \ \ \ \rho(0,y,t)=1.4,\ u(0,y,t)=0.1,\ v(0,y,t)=0,\ p(0,y,t)=1,\ y\in[0,1],t \in (0,2], &\\
    &\ \ \ \ \ \ \ \ \rho(1,y,t)=1.0,\ u(1,y,t)=0.1,\ v(1,y,t)=0,\ p(1,y,t)=1,\ y\in[0,1],t \in (0,2], &\\
    &\ \ \ \ \ \ \ \ \rho(x,0,t)=
    \begin{cases}
        1.4, & 0\le x\le0.5+0.1t, \\
        1.0, & 0.5+0.1t<x\le1,
    \end{cases} x\in[0,1],t\in(0,2], &\\
    &\ \ \ \ \ \ \ \ u(x,0,t)=0.1,\ v(x,0,t)=0,\ p(x,0,t)=1,\ x\in[0,1],t\in(0,2], &\\
    &\ \ \ \ \ \ \ \ \rho(x,1,t)=
    \begin{cases}
        1.4, & 0\le x\le0.5+0.1t, \\
        1.0, & 0.5+0.1t<x\le1,
    \end{cases} x\in[0,1],t\in(0,2], &\\
    &\ \ \ \ \ \ \ \ u(x,1,t)=0.1,\ v(x,1,t)=0,\ p(x,1,t)=1,\ x\in[0,1],t\in(0,2].&
\end{flalign*}
The solution for this problem is given by
\begin{flalign*}
    &\ \ \ \ \ \ \ \ \rho(x,y,t)=
    \begin{cases}
        1.4, & 0\le x\le0.5+0.1t,\ 0\le y\le1, t \in [0,2]\\
        1.0, & 0.5+0.1t<x\le1,\ 0\le y\le1, , t \in [0,2]
    \end{cases} &\\
    &\ \ \ \ \ \ \ \ u(x,y,t)=0.1,\ v(x,y,t)=0,\ p(x,y,t)=1, (x,y)\in[0,1]^2, t\in[0,2]. &
\end{flalign*}
In fact, we employ the exact solution values of variables $\rho$, $u$, $v$, and $p$ at the initial time as the initial values for the system, and the exact solution values of these variables at the geometric boundaries serve as the Dirichlet boundary values.

\begin{figure}[h!]
    \centering
    \begin{subfigure}{0.24\textwidth}
        \includegraphics[width=\textwidth]{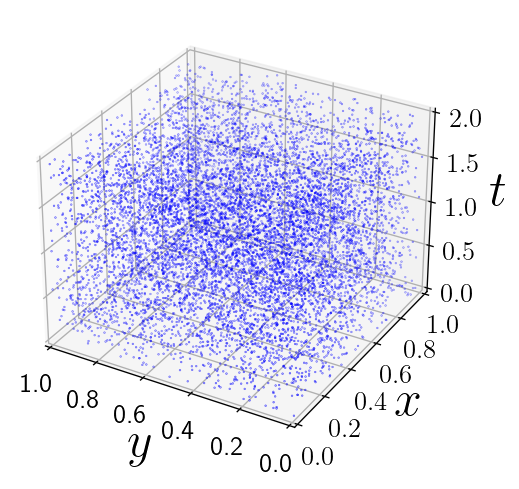}
        \caption{}
    \end{subfigure}
    \begin{subfigure}{0.24\textwidth}
        \includegraphics[width=\textwidth]{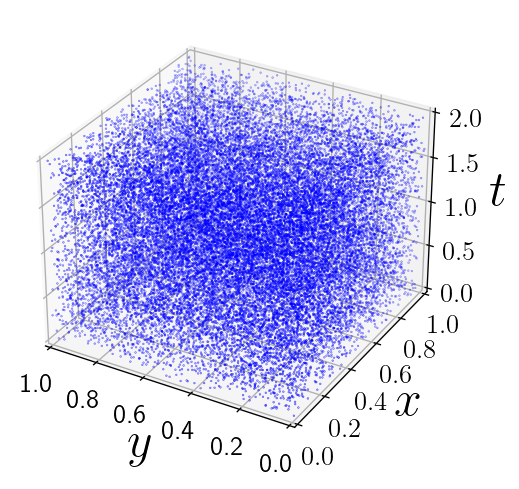}
        \caption{}
    \end{subfigure}
    \begin{subfigure}{0.24\textwidth}
        \includegraphics[width=\textwidth]{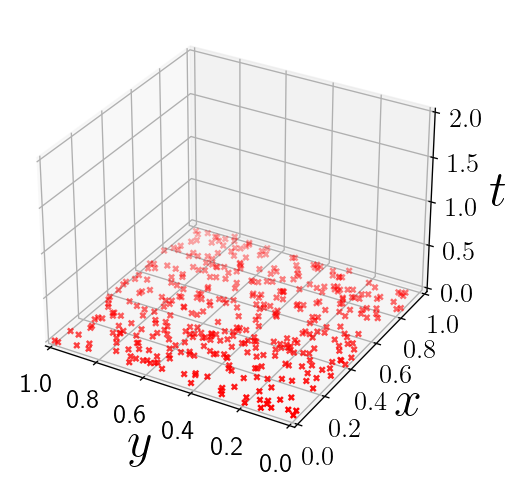}
        \caption{}
    \end{subfigure}
    \begin{subfigure}{0.24\textwidth}
        \includegraphics[width=\textwidth]{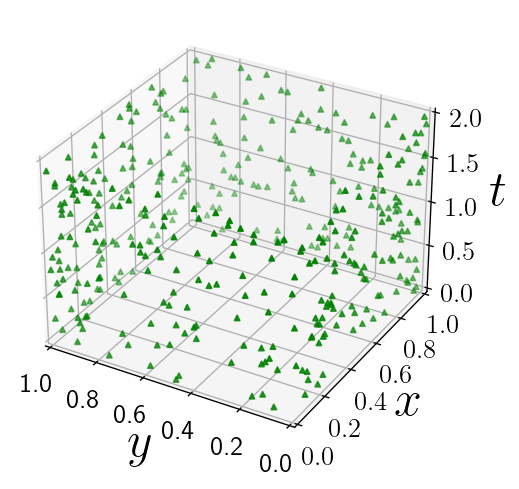}
        \caption{}
    \end{subfigure}
    \caption{Distributions of 10,000 (a) and 30,000 (b) training points corresponding to the governing equations and training points corresponding to the initial conditions (c) and boundary conditions (d) for solving the two-dimensional Euler equation.}
    \label{fig:euler2_point}
\end{figure}

We conduct a comparative analysis between two neural network architectures: ``5*256'' and ``256-16'' corresponding to PINNs and BsPINNs, respectively.
Within the neural network architecture, the $\tanh$ activation function is utilized in all hidden layers, except for the final layer, where a linear activation function is employed.
For training points corresponding to the governing equations, we randomly sample $10,000$ and $30,000$ training points over the domain $(0, 1)^2\times(0, 2]$, respectively.
In the subsequent text, we shall refer to these two cases as employing $10,000$ and $30,000$ internal training points, respectively.
For training points corresponding to the initial conditions, we randomly sample $400$ points in $[0, 1]^2\times\{0\}$.
Regarding training points for the boundary conditions, we randomly sample $100$ points for each edge of the $xy$-plane, totaling of $400$ training points for the boundary conditions.
The visual representation of these training points is illustrated in Fig.\ \ref{fig:euler2_point}, with blue, red, and green points corresponding to the governing equations, initial conditions, and boundary conditions respectively.
Following the notations introduced in Section \ref{s2}, we set ${\lambda}_I={\lambda}_B=1$ in the loss function, which remains consistent with the configuration in \cite{mao2020physics}.
We fix the number of iterations to $5,000$ and set an initial learning rate of $0.001$.

\begin{figure}[h!]
    \centering
    \includegraphics[width=.3\textwidth]{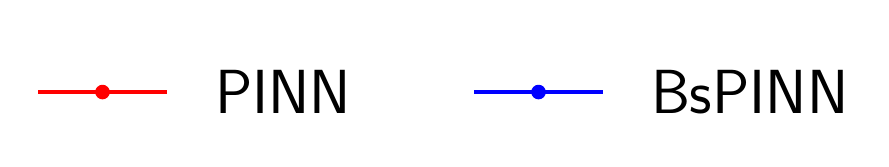} \\
    \includegraphics[width=0.4\textwidth]{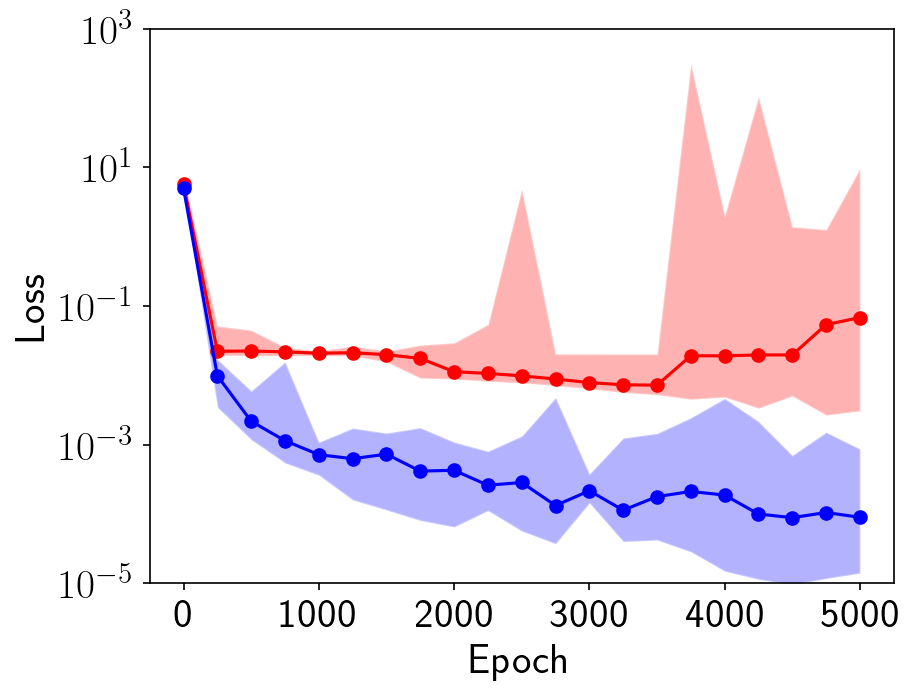}
    \includegraphics[width=0.4\textwidth]{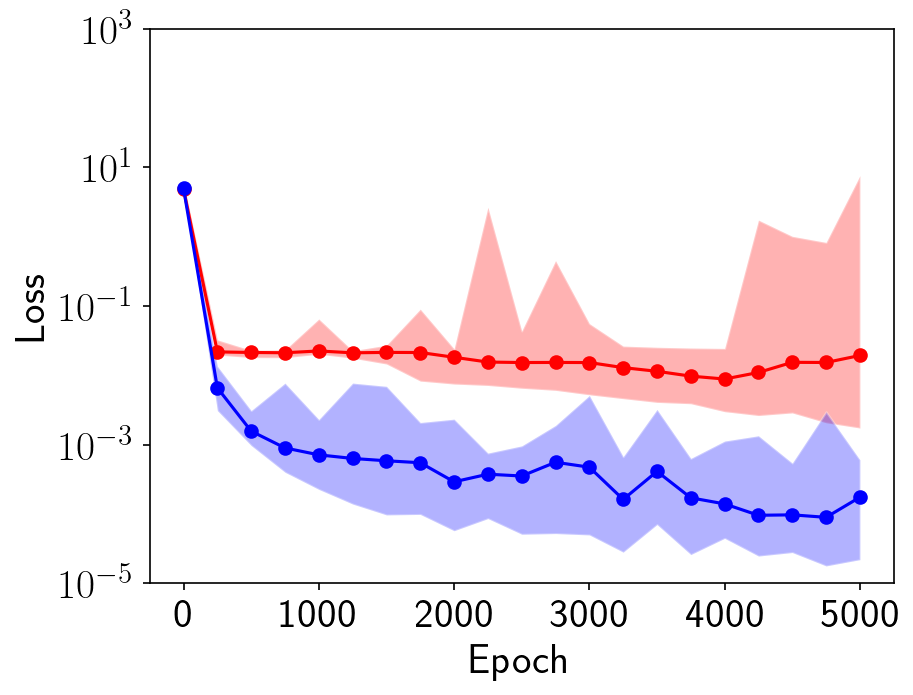}
    \caption{The dynamic change curves of the training loss for the PINNs and BsPINNs when solving the two-dimensional Euler equation with $10,000$ (left) and $30,000$ (right) interior training points. The shaded region here represents the segment between the maximum and minimum values of the loss at specific iteration steps from $10$ runs with different seeds, while the circular nodes represent the median values.}
    \label{fig:euler2_loss}
\end{figure}

Fig.\ \ref{fig:euler2_loss} depicts the dynamic change in the loss values concerning the training steps.
For both scenarios involving $10,000$ and $30,000$ interior training points, the median loss-value curves of the PINNs shows a noticeable plateau around the value of approximately $0.02$ in the early stages of training.
Around $1,500$ epochs, the median loss-value curves of the PINNs exhibit a gradual decrease, although they consistently remain significantly higher than those of BsPINNs at specific iterations.
In stark contrast, for both scenarios of $10,000$ and $30,000$ interior training points, the median loss-value curves of the BsPINNs steadily and consistently decrease without reaching a plateau at any epoch. This demonstrates a remarkable advantage of BsPINNs over PINNs in terms of convergence speed.

During the error assessment, the spatio-temporal domain $[0,1]^2 \times [0,2]$ is divided into a uniform grid of $100 \times 100 \times 100$, involving subdividing the $xy$-plane into a uniform grid of $100 \times 100$ and dividing the temporal domain $[0,2]$ into a uniform grid using $100$ nodes.
From this $100 \times 100 \times 100$ grid, we extract $10^6$ grid nodes denoted as $(x_i, y_i, t_i)$, where $i = 1, 2, ..., N$ and $N = 10^6$.
Subsequently, the relative error is calculated as
\begin{flalign*}
    &\ \ \ \ \ \ \ \ \text{error} = \frac{\sqrt{\sum_{i=1}^{N}\left(u_{\theta}\left(x_i,y_i,t_i\right)-u^\ast\left(x_i,y_i,t_i\right)\right)^2}}{\sqrt{\sum_{i=1}^{N}\left(u^\ast\left(x_i,y_i,t_i\right)\right)^2}},&
\end{flalign*}
where $u_{\theta}$ represents the predicted solution computed by neural network parameterized by $\theta$, and $u^\ast$ denotes the reference solution.
For the scenario involving $10,000$ interior training points, the calculated relative error for the BsPINNs is $1.668 \times 10^{-2} \pm 3.678 \times 10^{-3}$, while for the PINNs, it measures $4.751 \times 10^{-2} \pm 1.923 \times 10^{-2}$.
In the case involving $30,000$ interior training points, the BsPINNs maintain their excellence with a relative error of $1.443 \times 10^{-2} \pm 3.084 \times 10^{-3}$. In contrast, the PINNs exhibit a relative error of $5.044 \times 10^{-2} \pm 2.092 \times 10^{-2}$. Also, the standard deviation of the relative errors for the BsPINNs is approximately an order of magnitude smaller than that of the PINNs, demonstrating the stability of BsPINNs' performance.

\begin{figure}[h!]
    \centering
    \includegraphics[width=.3\textwidth]{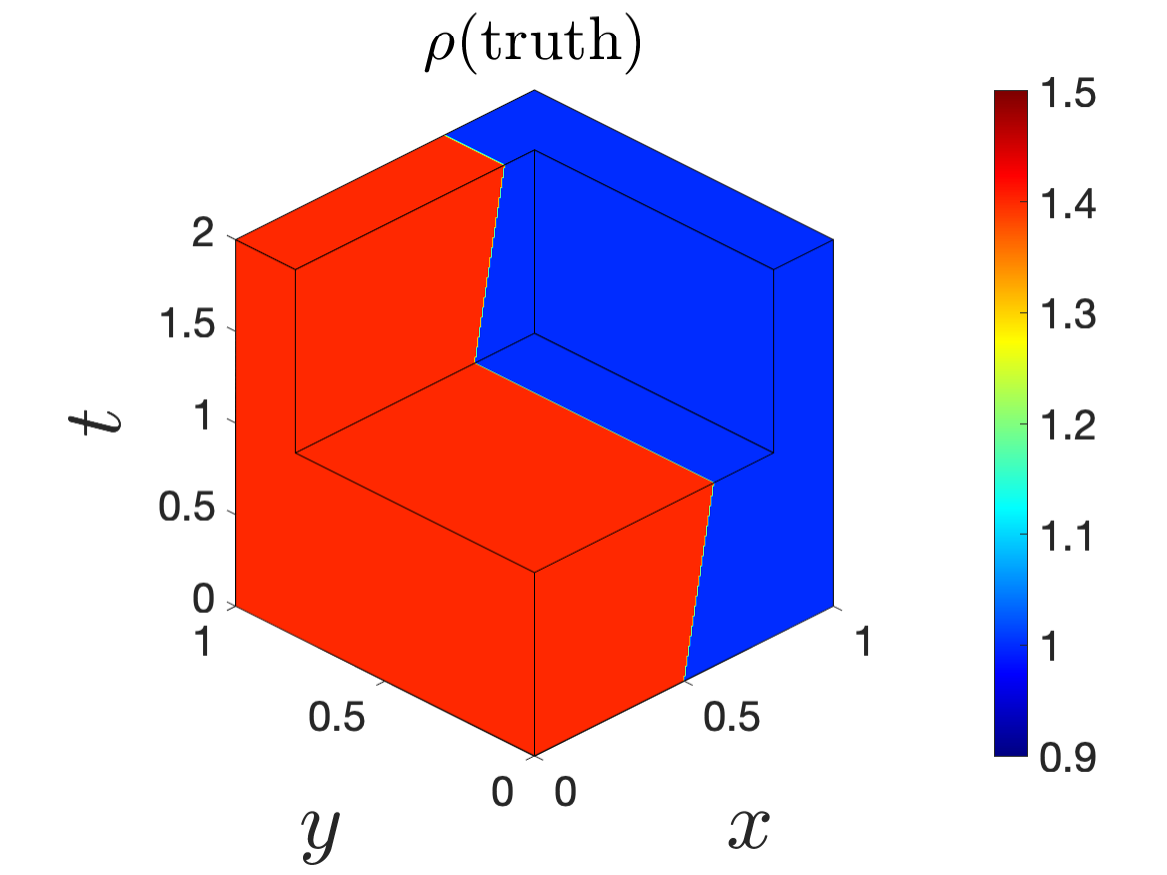}
    \includegraphics[width=.3\textwidth]{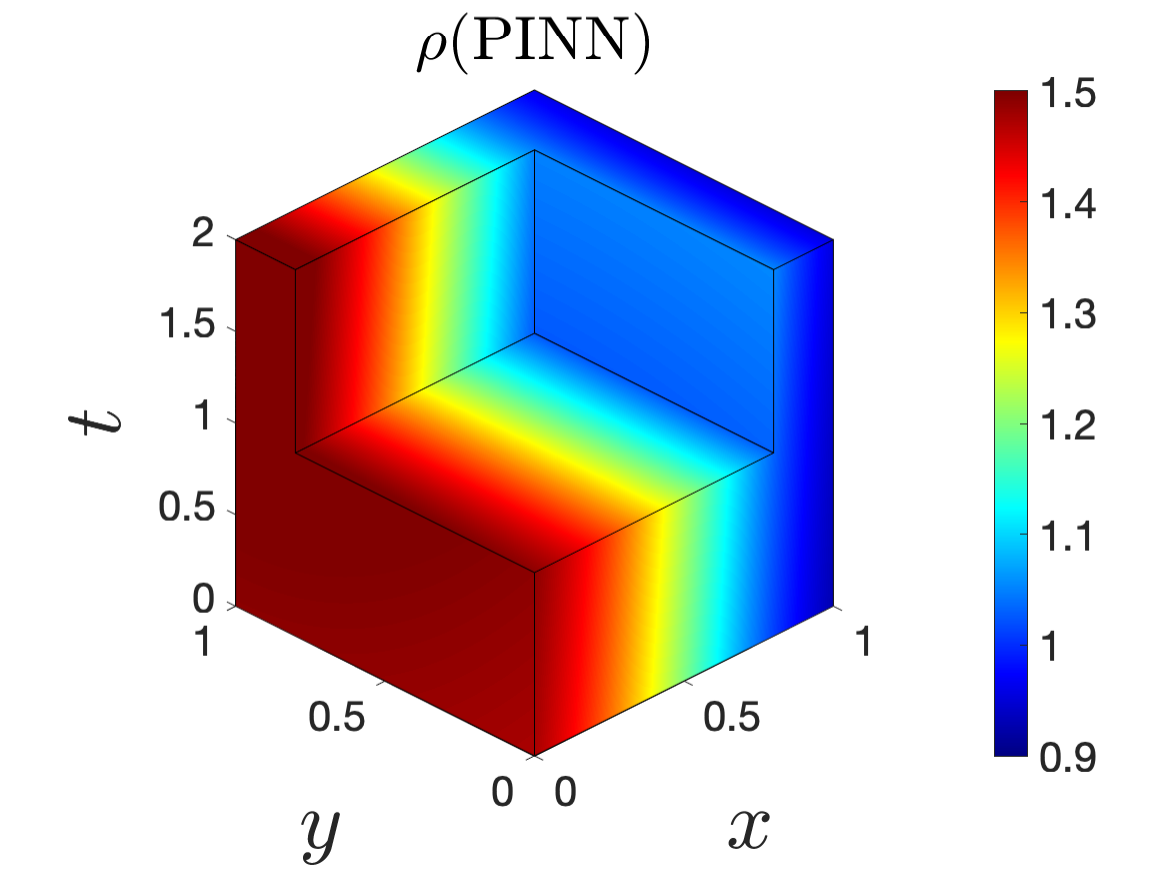}
    \includegraphics[width=.3\textwidth]{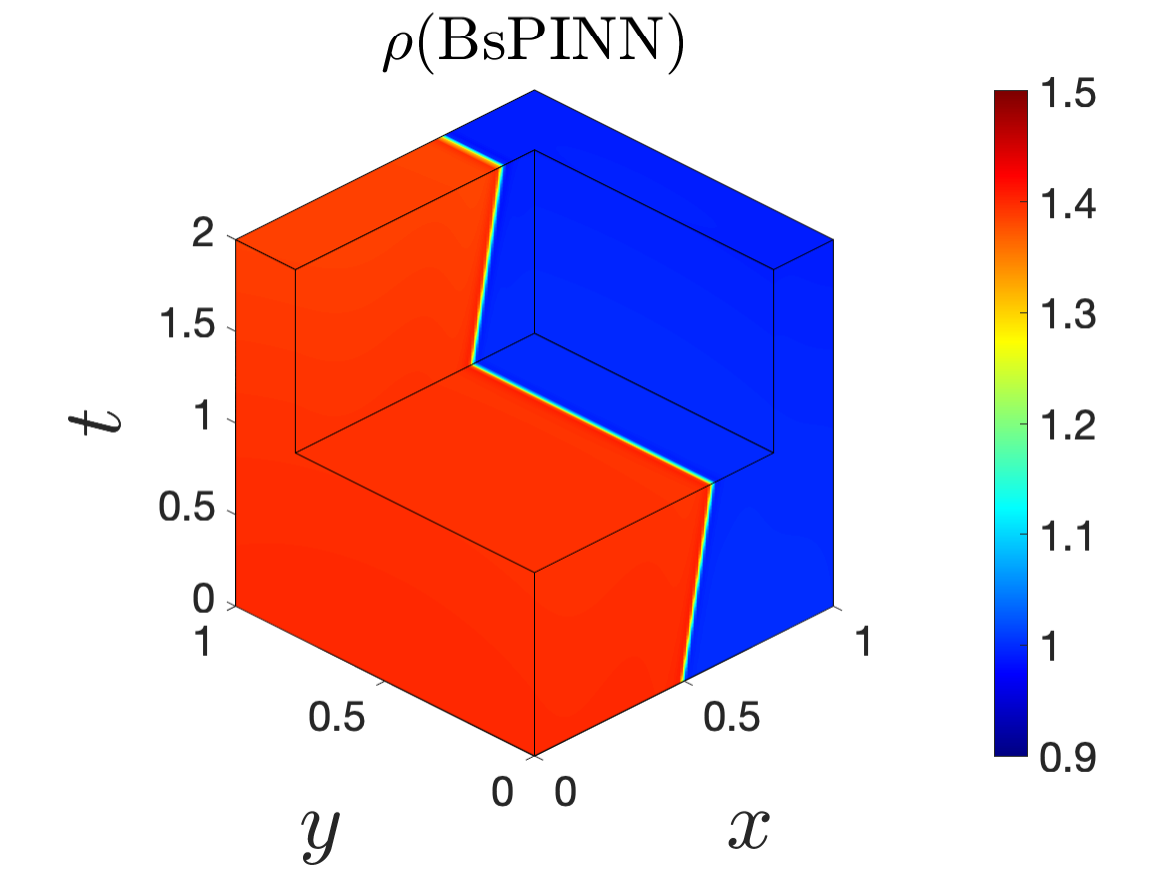} \\
    \includegraphics[width=.3\textwidth]{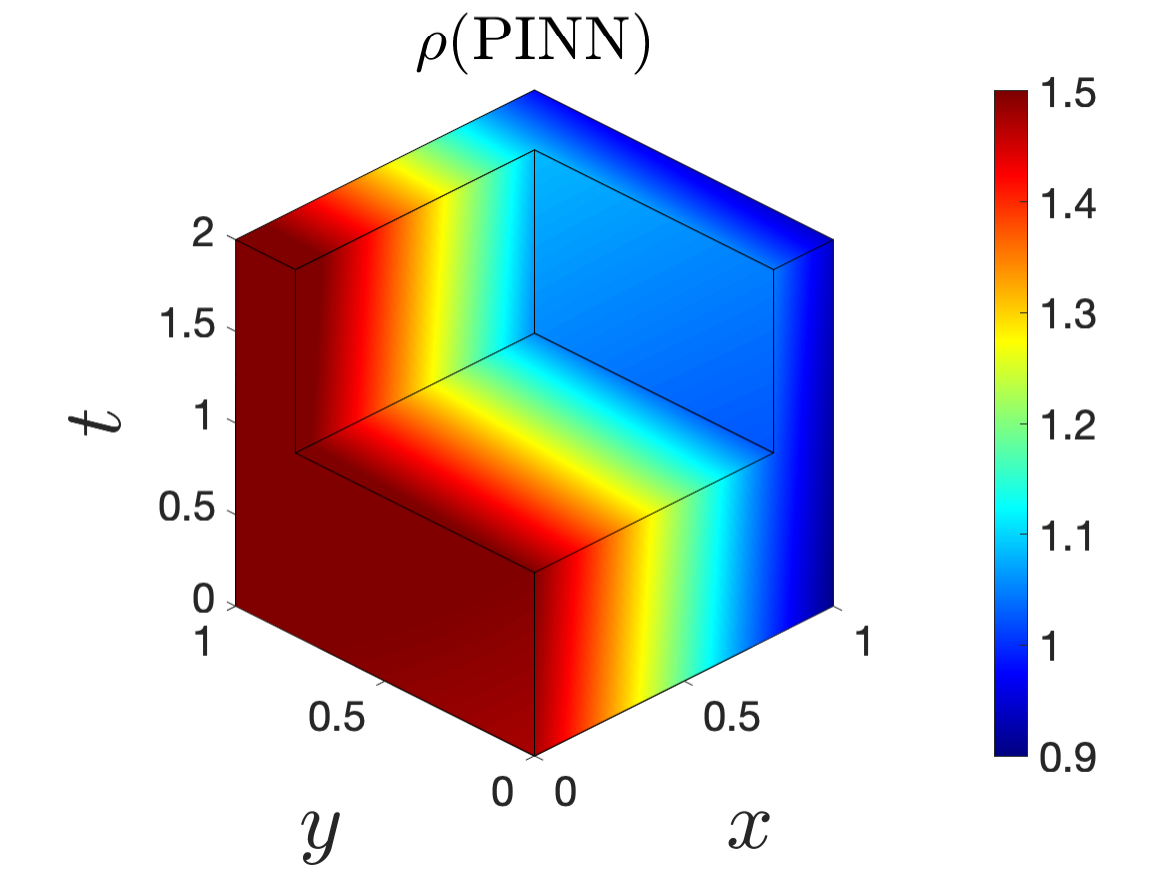}
    \includegraphics[width=.3\textwidth]{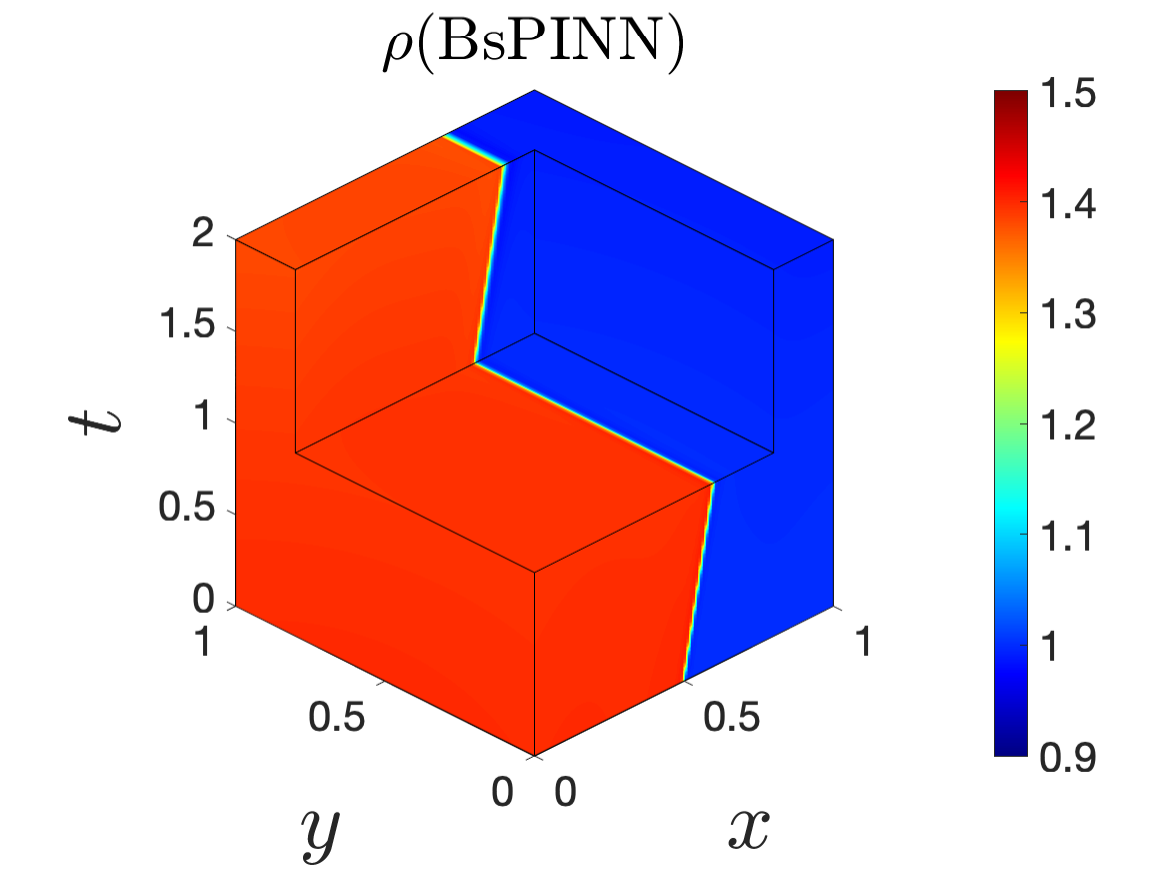}
    \caption{Top: Exact solution(left) and predicted solutions of PINN(middle) and BsPINN(right) in terms of the density $\rho$ for the two-dimensional Euler equation with $10,000$ interior training points. Bottom: The predicted solutions of PINN(left) and BsPINN(right) in terms of the density $\rho$ for the two-dimensional Euler equation with $30,000$ interior training points.}
    \label{fig:euler2_r}
\end{figure}

In Fig.\ \ref{fig:euler2_r}, we display predicted solutions for the PINNs with $10,000$ and $30,000$ interior training points, along with corresponding solutions for the BsPINNs at the same training point counts. Each prediction comes from the run with the minimum relative error out of $10$.
From Fig.\ \ref{fig:euler2_r},  it is evident that for both cases with $10,000$ and $30,000$ interior training points, the outputs of the BsPINNs more accurately approximates the regions of numerical discontinuity, whereas the outputs of the PINNs only captures a vague semblance of these regions. In other words, BsPINNs demonstrate better capabilities in approximating solutions with discontinuous regions compared to PINNs.

Furthermore, as depicted in Fig.\ \ref{fig:euler2_r}, the PINN exhibits a larger relative error when utilizing $30,000$ interior training points as opposed to $10,000$. To elucidate this difference, we analyze the phenomenon through the lens of transition points \cite{shen2014improvement} and the concept of a long-tailed distribution \cite{resnick2007heavy}.

Transition points, as described in \cite{shen2014improvement}, are those that link smooth regions and discontinuities, significantly impacting the solution accuracy near these discontinuities.
Guided by this concept, we refer to the training points near the discontinuities during the neural network training process as ``transition points'', emphasizing their pivotal role in capturing the discontinuity.
The method for selecting transition points in this experiment involves conceptualizing the spatio-temporal solution domain as a three-dimensional spatio space, dividing it into a uniform hexahedral element grid. For a set of $n$ training points, we establish $m = \lfloor \sqrt[3]{n} \rfloor$, partitioning the spatio-temporal solution domain into a uniform grid of $m \times m \times m$.

\begin{figure}[h!]
    \centering
    \includegraphics[width=.3\textwidth]{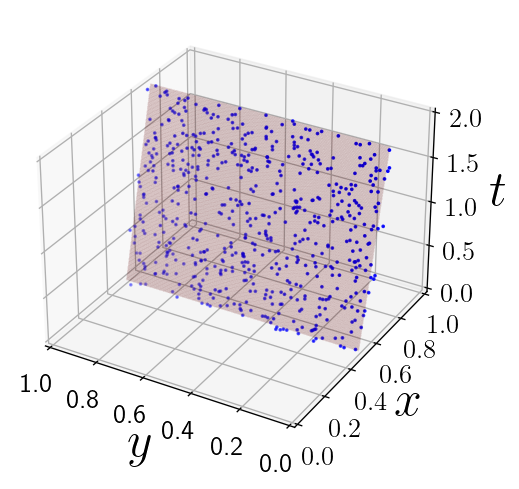}
    \includegraphics[width=.3\textwidth]{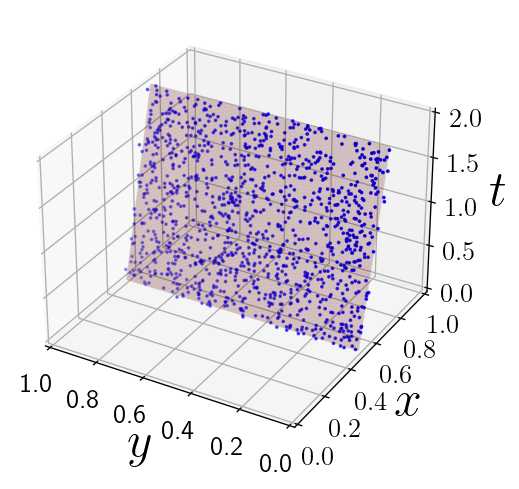}
    \caption{Schematic of the transition points in the case of $10,000$ (left) and $30,000$ (right) interior points, containing $311$ and $603$ points respectively. The proportions of training points to the total interior training points in these two cases are $6.09\%$ and $4.02\%$, respectively. Due to the scarcity of points here, individual point sizes have been increased compared to Fig.\ \ref{fig:euler2_point}}
    \label{fig:euler2_interface}
\end{figure}

In Fig.\ \ref{fig:euler2_interface}, the transition points are displayed in the case of $10,000$ and $30,000$ interior training points, containing $609$ and $1205$ points respectively.
Here, the transition points represent $6.09\%$ and $4.02\%$ of the interior training points, showcasing that transition points constitute only a small fraction of the interior training points.

Considering the perspective of the long-tailed distribution \cite{resnick2007heavy}, transition points can be labeled as ``tail category" points, while other points could be referred to as ``head category" points. In the study \cite{tan2020equalization}, it is noted that as the proportion of head category points in a long-tailed distribution increases, reducing the loss for tail category points becomes more challenging. This challenge highlights the difficulty in achieving high accuracy with PINNs. The relatively lower proportion of transition points in the scenario using $30,000$ internal training points might contribute to the lower accuracy observed in PINNs.

In contrast, the accuracy of the BsPINN when trained with $30,000$ interior points surpasses that of training with $10,000$ interior training points.
This indicates that BsPINNs utilize multi-channel structure of BsNNs, allowing each channel to focus on features within a specific ``local region'', within which the proportion of transition points acting as tail category training points does not decrease, despite the reduced overall proportion of transition points across the entire solution domain.
Consequently, within these local regions, the proportion of the tail category  training points does not decrease and the total number of interior training points increases. This allows BsPINNs to achieve higher accuracy in the scenario with $30,000$ interior training points compared to the situation with $10,000$ interior training points.

By drawing these comparisons, the results from both experiments further underscore the efficacy of BsPINNs in accurately addressing PDEs with discontinuous solutions.

\subsection{Two-dimensional Helmholtz equation}\label{s_highhe}
Within this subsection, the Helmholtz equation, as outlined in \cite{Escapil2022}, within a two-dimensional square domain $[0,1]^2$, is considered.
The expression of the two-dimensional Helmholtz equation is
\begin{flalign*}
    &\ \ \ \ \ \ \ \ -\Delta u - \kappa^{2} u = f \text { in } \Omega, &\\
    &\ \ \ \ \ \ \ \ u = 0 \text { on } \partial\Omega,&
\end{flalign*}
where $\Omega=(0,1)^2$, $\kappa$ represents the wavenumber, which is set to be $8\pi$ in this experiment, $u$ represents the displacement, and $f$ represents the external force, which is set to be $f(x,y)=\kappa^2\sin(\kappa x)\sin(\kappa y)$ in this case. The equation admits the exact solution
\begin{flalign*}
    &\ \ \ \ \ \ \ \ u^\ast\left(x,y\right)=\sin(\kappa x)\sin(\kappa y).&
\end{flalign*}

We compare the performance of two distinct neural network architectures: ``5*256" for PINNs and ``256-16" for BsPINNs.
Within these architectures, the sine activation function is deployed across all hidden layers, while the final layer employs a linear activation function.
For training points aligned with the governing equation, we randomly select $6561$ points within $\Omega$. Additionally, for training points related to the boundary condition, we uniformly sample $80$ points from each edge of $\Omega$, totaling $320$ training points for the boundary condition.
Rereferring to the symbols introduced in section \ref{s2}, we set ${\lambda}_B=100$ in the loss function.
Our training process spans a total of $80,000$ epochs, with an initial learning rate of $0.001$.

\begin{figure}[h!]
    \centering
    \includegraphics[width=.4\textwidth]{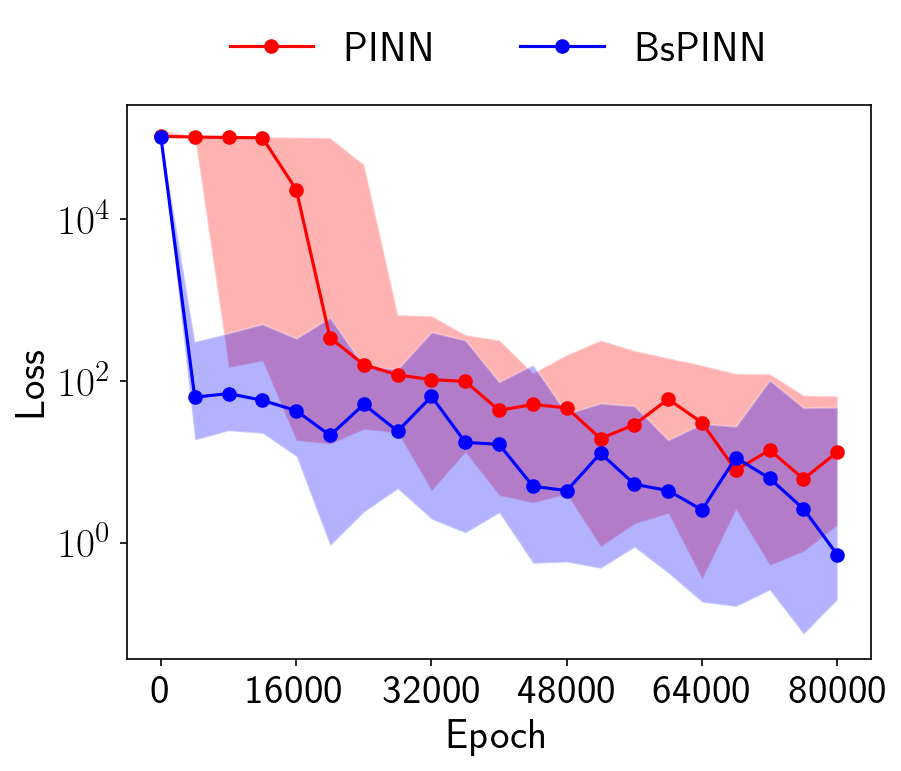}
    \caption{The dynamic change curves of the training loss for the PINNs and BsPINNs when solving the two-dimensional Helmholtz equation. The shaded regions here represent the segment between the maximum and minimum values of the loss at specific iteration steps from $10$ runs with different seeds, while the circular nodes represent the median values.}
    \label{fig:highhe_loss}
\end{figure}

Fig.\ \ref{fig:highhe_loss} illustrates the changes in the loss values of the PINNs and BsPINNs.
A noticeable distinction between the loss values of the PINNs and BsPINNs emerges during the early stage of training. Specifically, the loss values of the PINNs hover around $10^5$, with a significant decrease observed only after approximately $15,000$ epochs. In contrast, BsPINNs do not encounter this issue, as their loss values rapidly decrease to below $100$, indicating a significant advantage in convergence speed over PINNs.
The average and standard deviation of the minimum training loss values for the PINNs from $10$ runs are $7.359 \times 10^{-1}$ and $5.550 \times 10^{-1}$, respectively. In contrast, for BsPINNs, the corresponding values are $1.489 \times 10^{-1}$ and $5.108 \times 10^{-2}$. This illustrates the advantage of BsPINNs in terms of both convergence and stability.

\begin{figure}[h!]
    \centering
    \includegraphics[width=.3\textwidth]{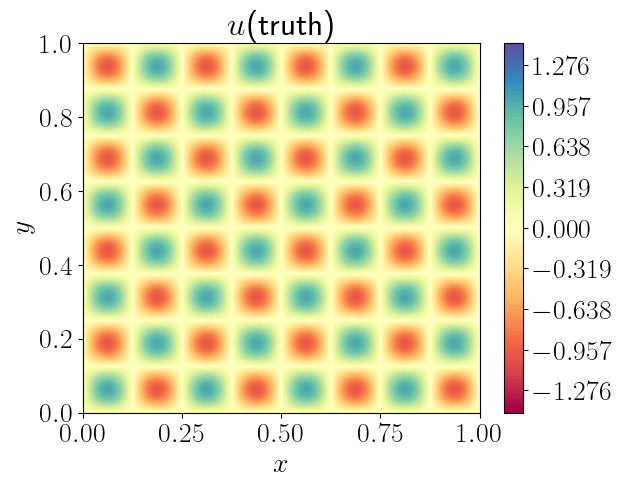}
    \includegraphics[width=.3\textwidth]{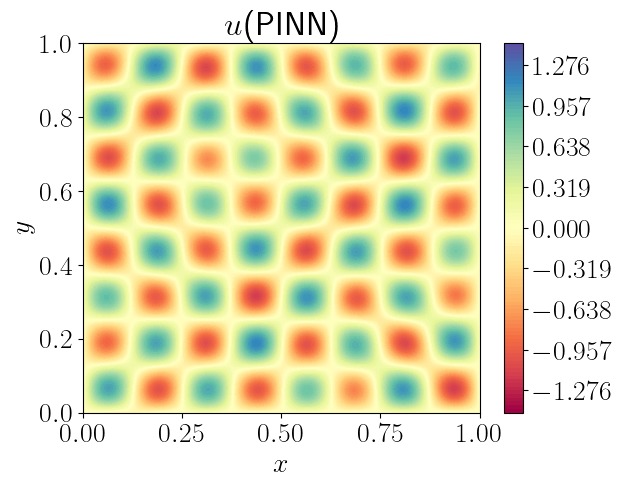} \\
    \includegraphics[width=.3\textwidth]{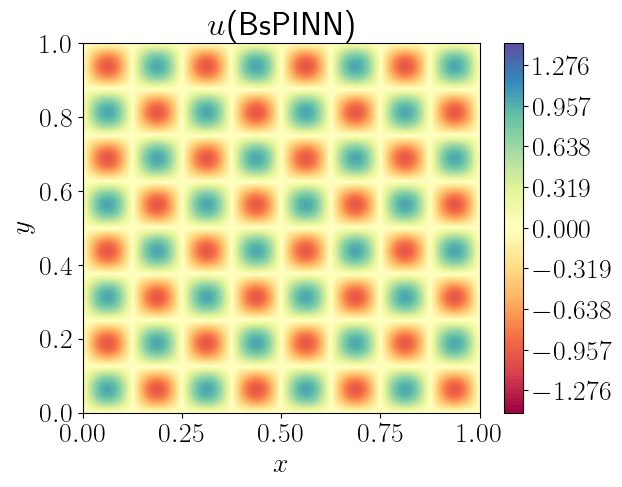}
    \caption{Top: Exact solution (left) and the predicted solution of the PINN (right) in terms of the displacement value $u$ for the two-dimensional Helmholtz equation. Bottom: The predicted solution of the BsPINN in terms of the displacement value $u$ for the two-dimensional Helmholtz equation.}
    \label{fig:highhe_u}
\end{figure}

In the process of error estimation, we partition the solution domain $[0,1]^2$ into a uniform grid of dimensions 500$\times$500 and extract the grid nodes denoted as $(x_i,y_i)(i=1,2,\ldots,N)$ where $N = 2.5 \times 10^5$. The relative error is calculated by
\begin{flalign*}
    &\ \ \ \ \ \ \ \ \text{error} = \frac{\sqrt{\sum_{i=1}^{N}\left(u_{\theta}\left(x_i,y_i\right)-u\left(x_i,y_i\right)\right)^2}}{\sqrt{\sum_{i=1}^{N}\left(u\left(x_i,y_i\right)\right)^2}},&
\end{flalign*}
where $u_{\theta}$ represents the predicted solution computed by neural network parameterized by $\theta$, and $u^\ast$ denotes the reference solution.
The relative error of the PINNs is $1.396 \times 10^{-1} \pm 2.940 \times 10^{-2}$, whereas for the BsPINNs,  it impressively decreases to $3.278 \times 10^{-2} \pm 1.900 \times 10^{-2}$.
In Fig.\ \ref{fig:highhe_u}, the reference solution is presented alongside the predicted solutions obtained from the PINN and BsPINN, each derived from the respective run with the minimum relative error out of $10$. Evidently, the predicted solution of the PINN exhibits noticeable errors in specific regions; for example, some valleys are not clearly separated. In contrast, the predicted solution of the BsPINN demonstrates a significantly closer alignment with the exact solution. This outcome unequivocally indicates that BsPINNs achieve notably higher accuracy compared to PINNs.

\begin{figure}[h!]
    \centering
    \includegraphics[width=0.24\textwidth]{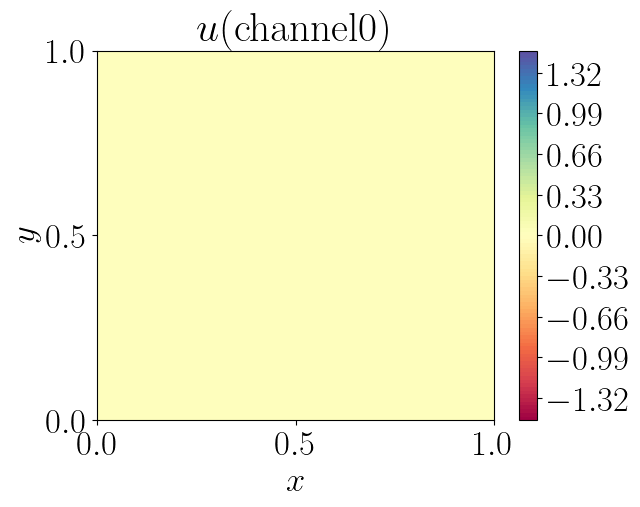}
    \includegraphics[width=0.24\textwidth]{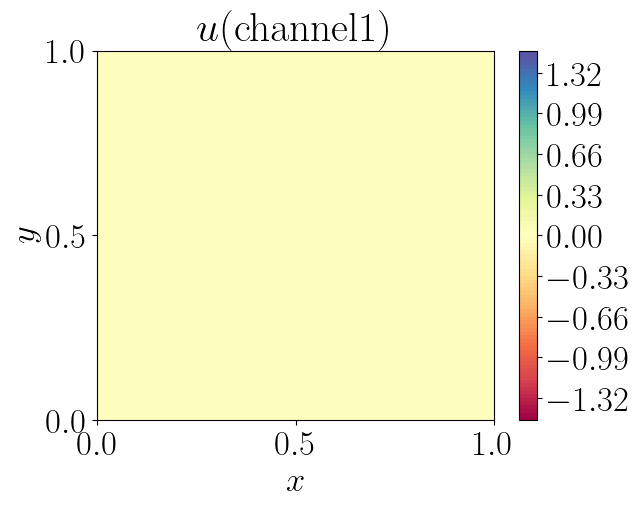}
    \includegraphics[width=0.24\textwidth]{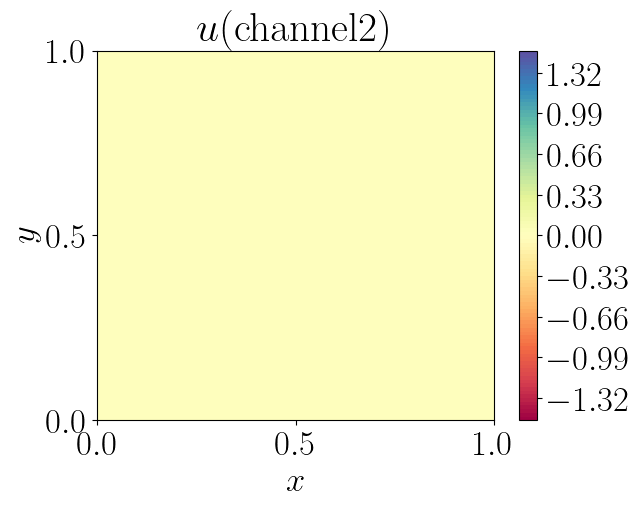}
    \includegraphics[width=0.24\textwidth]{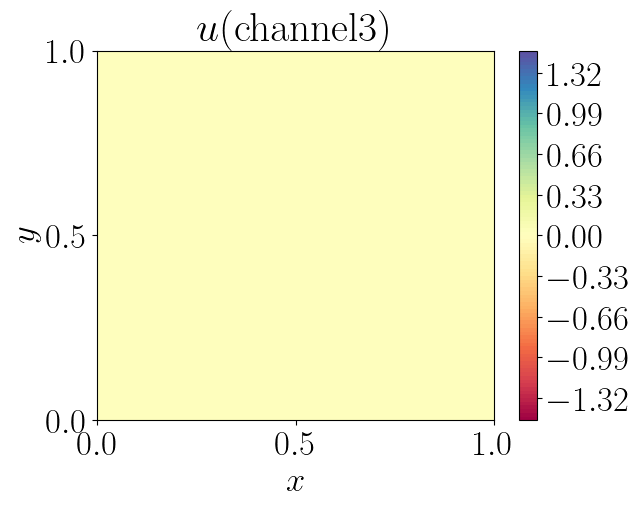} \\
    \includegraphics[width=0.24\textwidth]{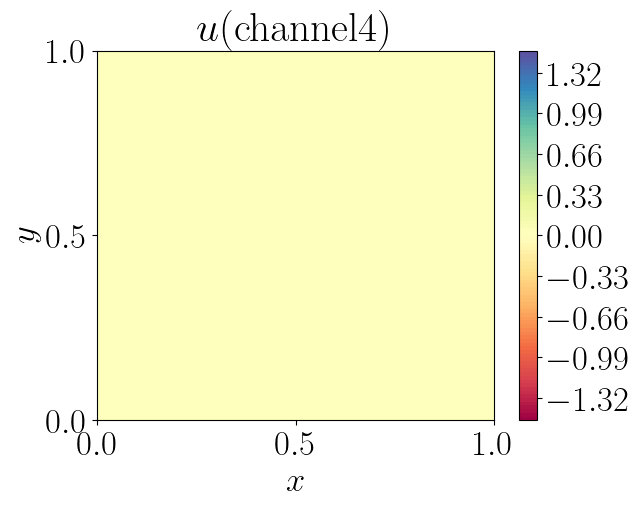}
    \includegraphics[width=0.24\textwidth]{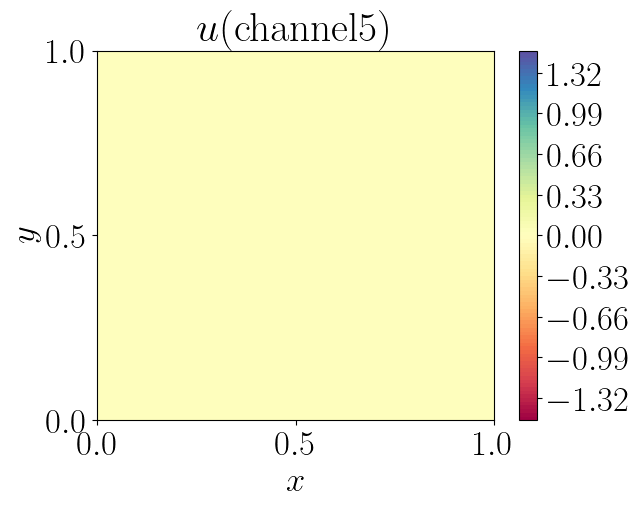}
    \includegraphics[width=0.24\textwidth]{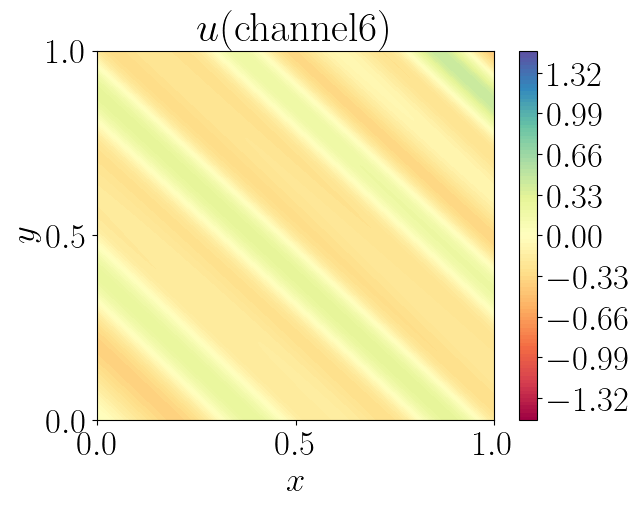}
    \includegraphics[width=0.24\textwidth]{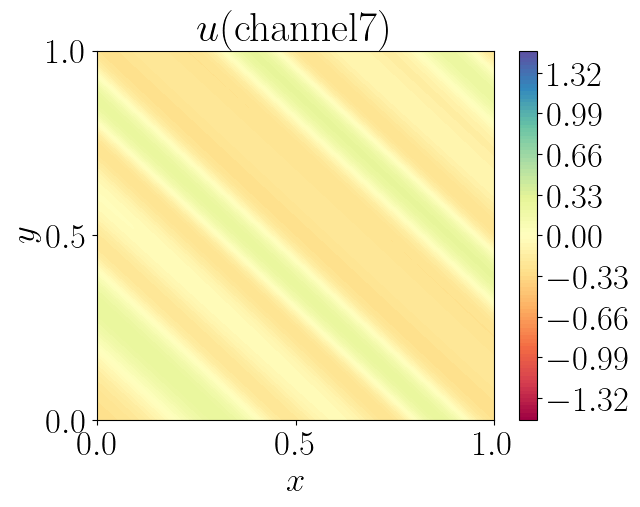} \\
    \includegraphics[width=0.24\textwidth]{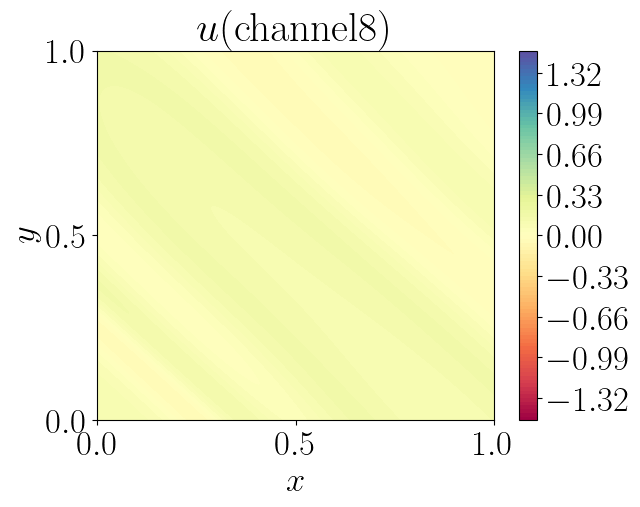}
    \includegraphics[width=0.24\textwidth]{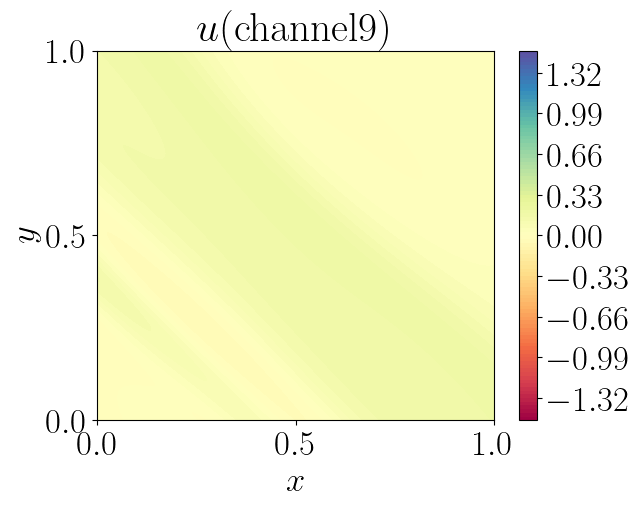}
    \includegraphics[width=0.24\textwidth]{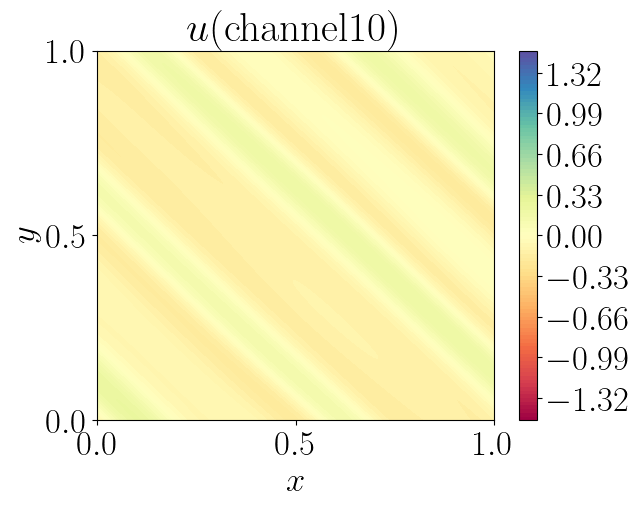}
    \includegraphics[width=0.24\textwidth]{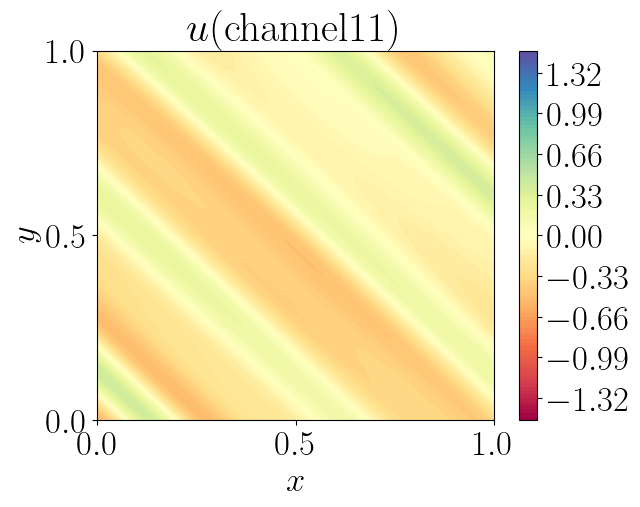} \\
    \includegraphics[width=0.24\textwidth]{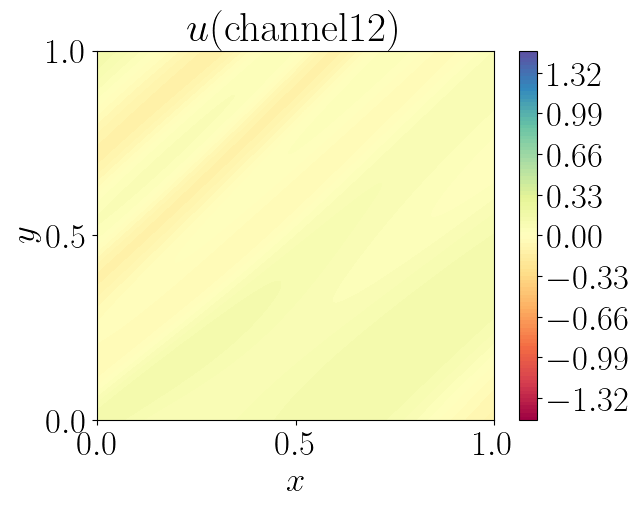}
    \includegraphics[width=0.24\textwidth]{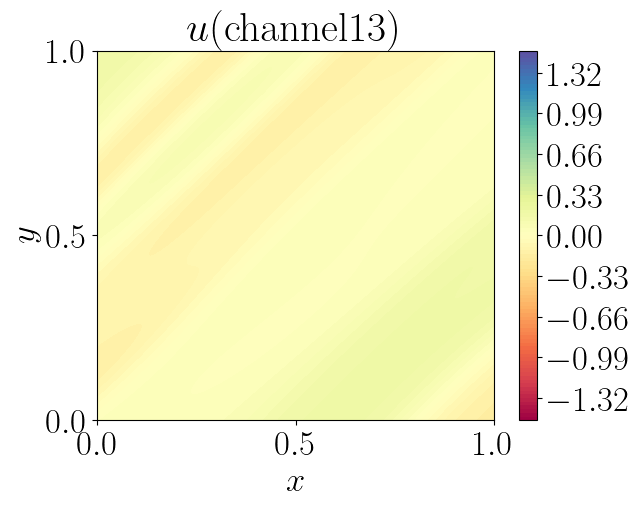}
    \includegraphics[width=0.24\textwidth]{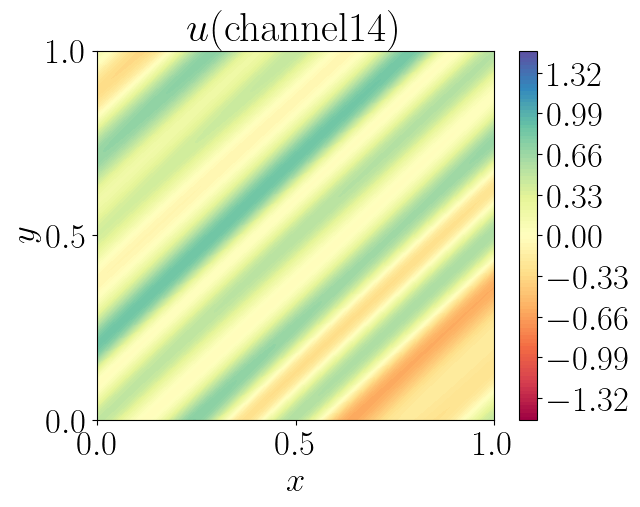}
    \includegraphics[width=0.24\textwidth]{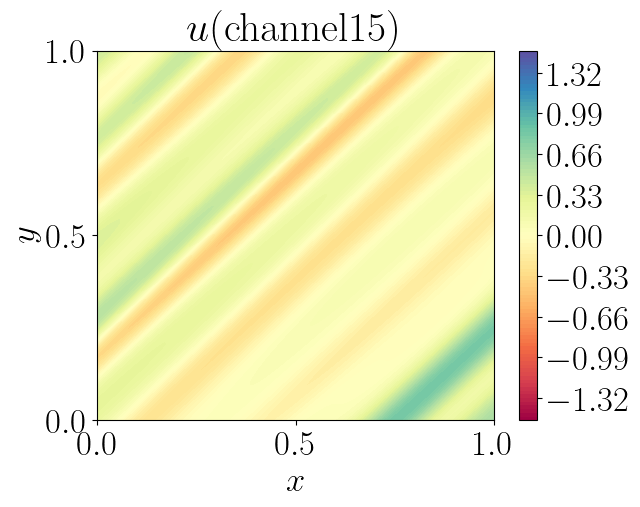}
    \caption{The images of predicted values for $16$ channels in the BsPINN used to solve the two-dimensional Helmholtz equation.}
    \label{fig:highhe_per_channel_BsPINN}
\end{figure}

\begin{figure}[h!]
    \centering
    \includegraphics[width=0.24\textwidth]{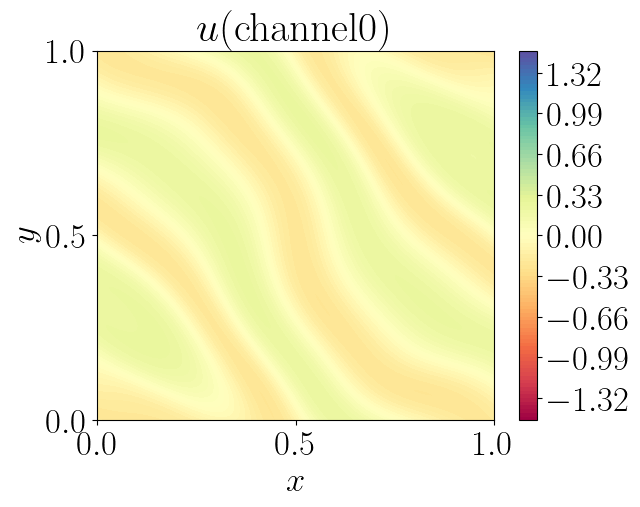}
    \includegraphics[width=0.24\textwidth]{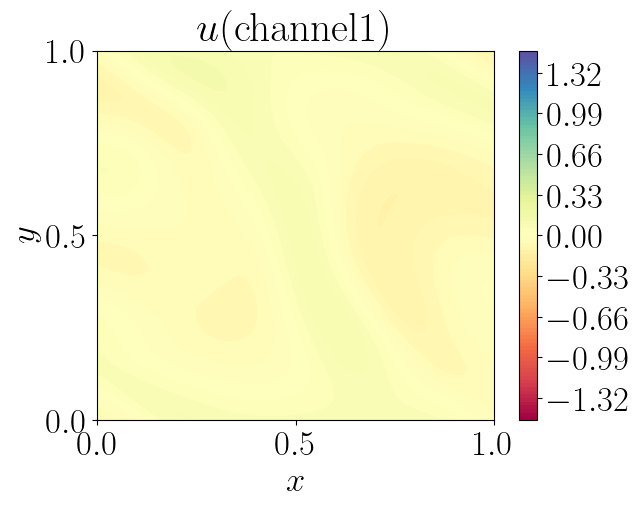}
    \includegraphics[width=0.24\textwidth]{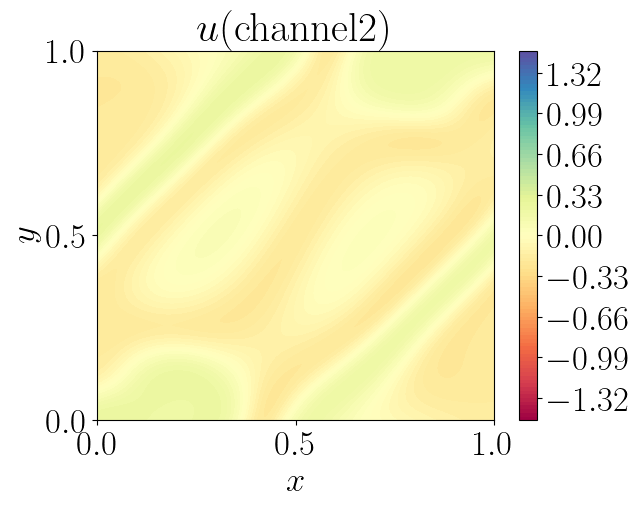}
    \includegraphics[width=0.24\textwidth]{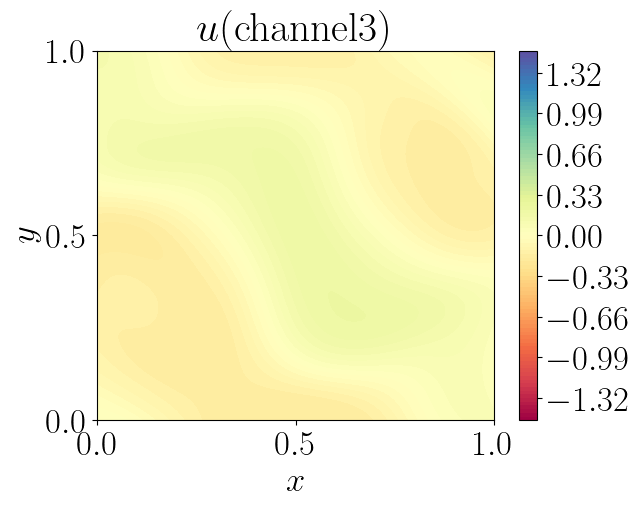} \\
    \includegraphics[width=0.24\textwidth]{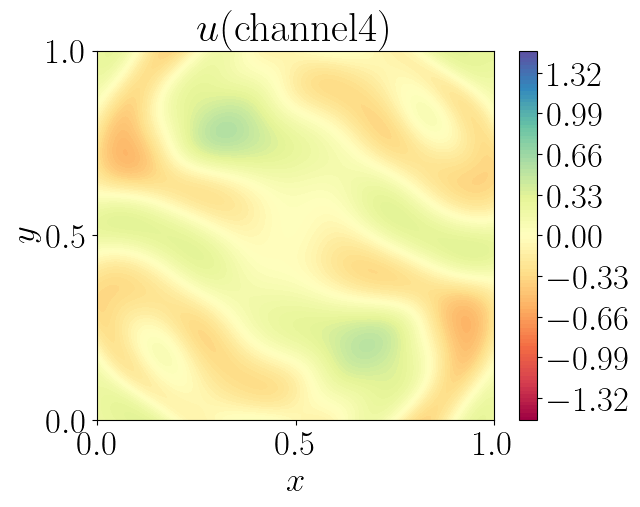}
    \includegraphics[width=0.24\textwidth]{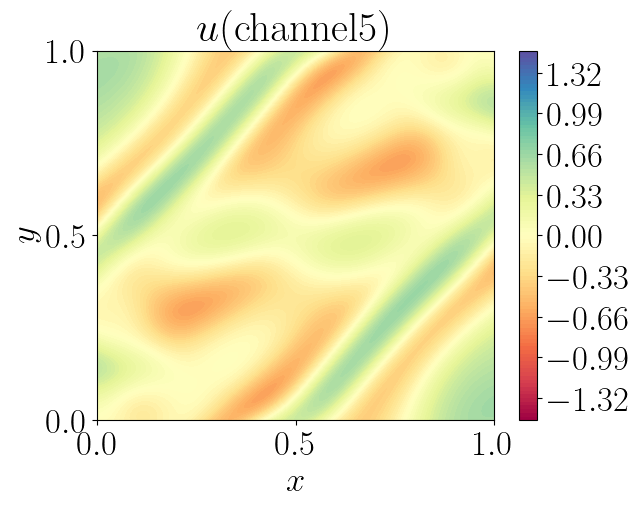}
    \includegraphics[width=0.24\textwidth]{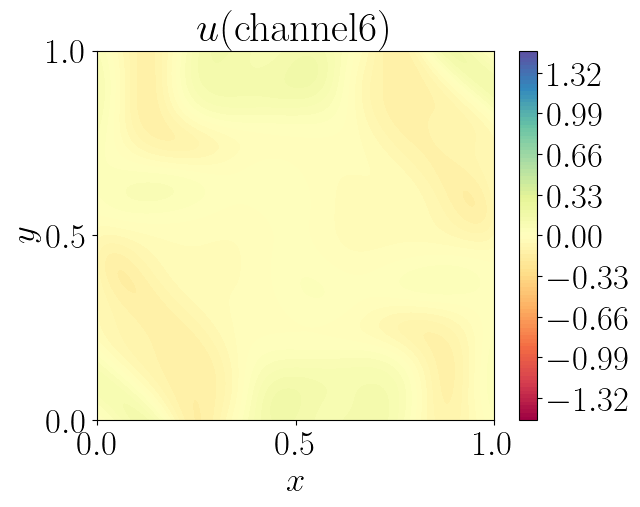}
    \includegraphics[width=0.24\textwidth]{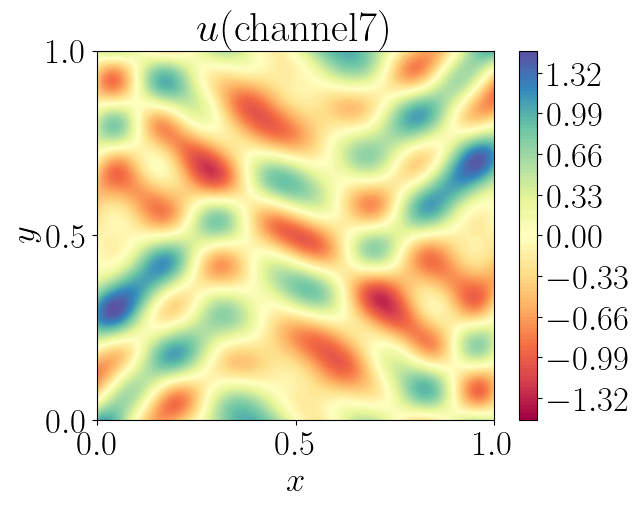} \\
    \includegraphics[width=0.24\textwidth]{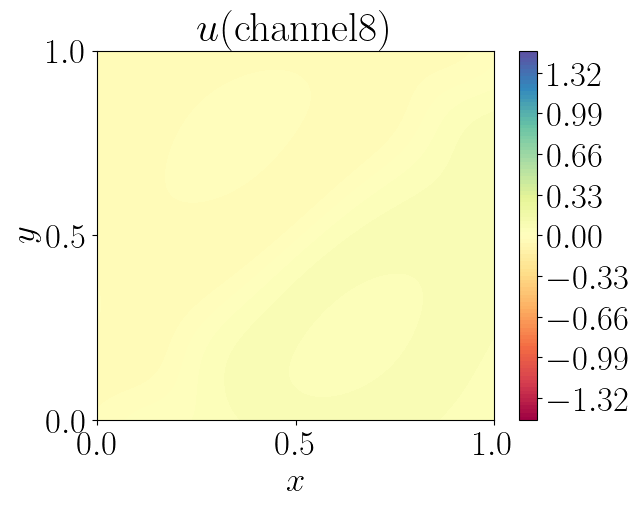}
    \includegraphics[width=0.24\textwidth]{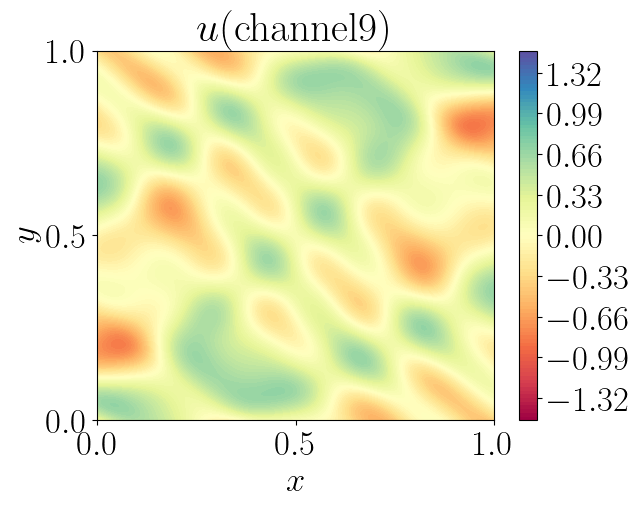}
    \includegraphics[width=0.24\textwidth]{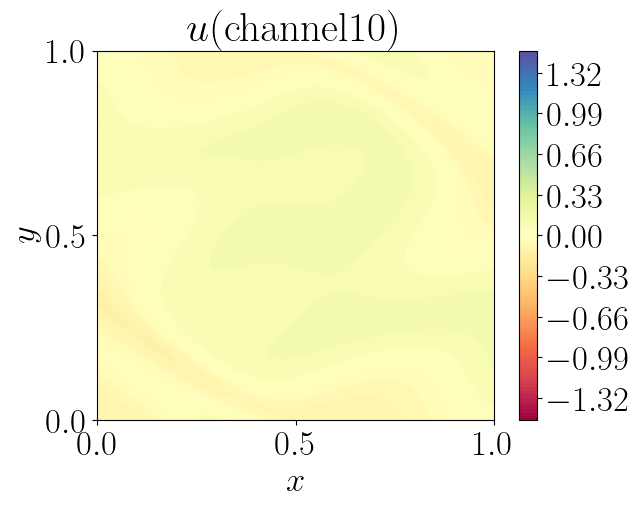}
    \includegraphics[width=0.24\textwidth]{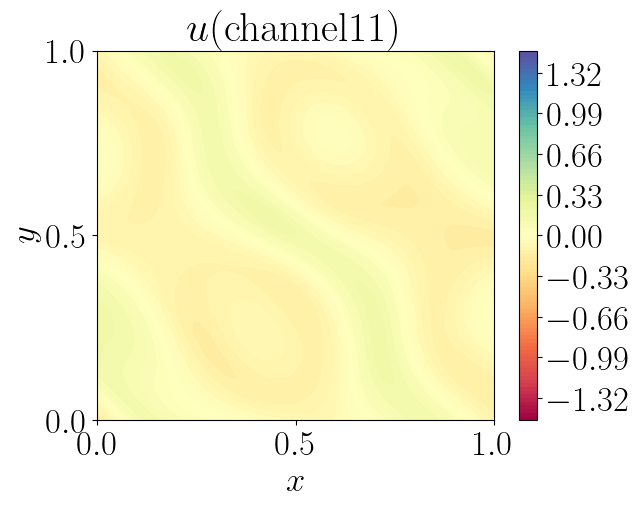} \\
    \includegraphics[width=0.24\textwidth]{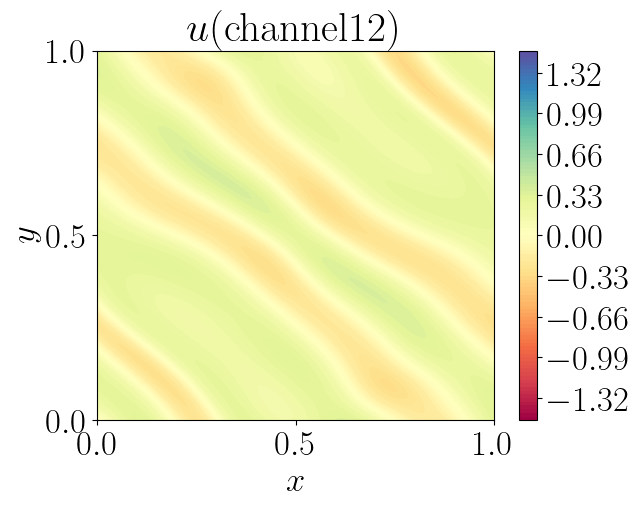}
    \includegraphics[width=0.24\textwidth]{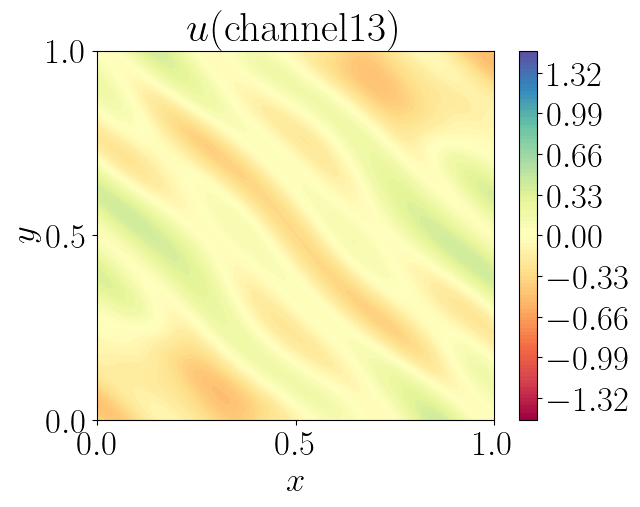}
    \includegraphics[width=0.24\textwidth]{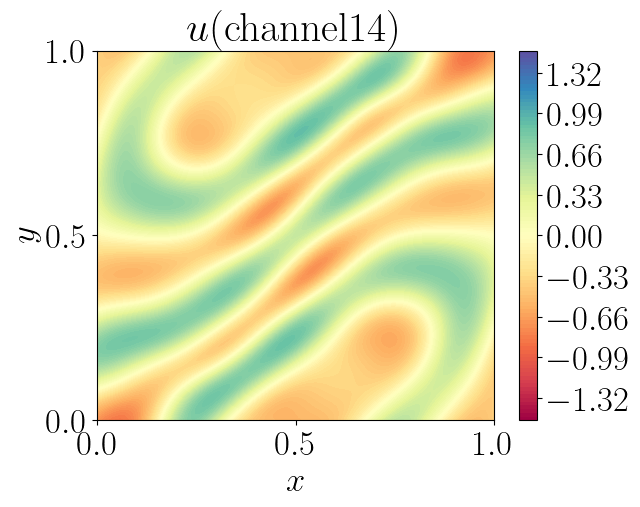}
    \includegraphics[width=0.24\textwidth]{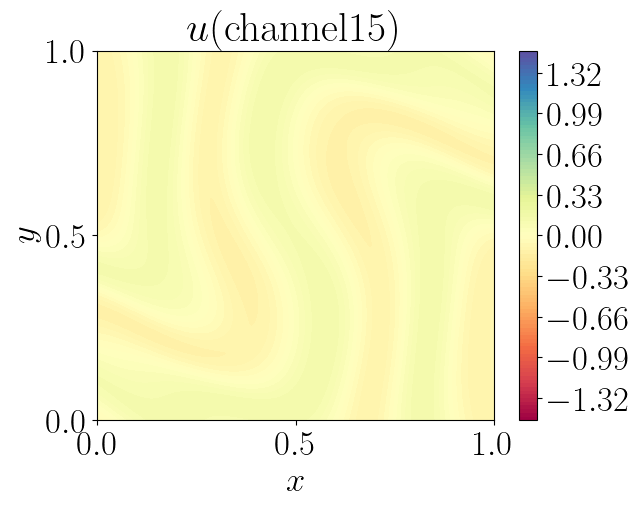}
    \caption{The images of predicted values for $16$ channels in the PINN used to solve the two-dimensional Helmholtz equation.}
    \label{fig:highhe_per_channel_pinn}
\end{figure}

To delve further into the reasons behind BsPINNs' performance, we examine the output of each channel in the BsPINN with the minimum relative error in $10$ runs. This assessment aims to determine whether BsPINNs comprehends ``local features", as stated in subsection \ref{s_motivation}.
We consider the output of each neuron block within the last hidden layer as the output of an individual channel in the BsPINN.
Fig.\ \ref{fig:highhe_per_channel_BsPINN} visualizes the outputs of each channel in the BsPINN.
Channels with outputs close to zero are indicative of their ineffectiveness or insignificance. Conversely, certain channels capture specific features of the exact solution projected onto a lower-dimensional space. We refer to these characteristics as the local features of the exact solution.
Essentially, due to the intricate nature of the exact solution, the BsPINN de-constructs it into a combination of multiple simpler functions, approximating each of these simpler functions through an individual channel.
Consequently, the summation of the outputs from all channels yields a result that closely mirrors the exact solution.
This ability of BsPINNs to decompose complex solutions into simpler components and represent them through distinct channels contributes to its remarkable accuracy in approximating the solution to the Helmholtz equation.

We further extract the output of every adjacent $16$ neurons in the last hidden layer of the PINN with the minimum relative error, considering it as the output of a channel.
As shown in Fig.\ \ref{fig:highhe_per_channel_pinn}, each channel in the PINN also exhibits a similar trend of learning local features as seen in the BsPINN (shown in Fig. \ref{fig:highhe_per_channel_BsPINN}), but it's not as pronounced.
In the PINN, which utilizes an FNN structure, each channel in the final hidden layer is connected to all neurons in the preceding layer. Consequently, when there is a change in the output of a particular channel, this alteration immediately affects the outputs of all channels through the neurons in the penultimate hidden layer, acting as intermediaries.
In the BsPINN, because the last few hidden layers are not fully connected, a change in the output of a specific channel doesn't easily affect the outputs of all other channels through an intermediate neuron.
The independence of each channel in the BsPINN allows them to specialize in learning different local features, thereby enabling the BsPINN to outperform the PINN.
%%% 删掉了下面这句话
%It is the difference in the interdependence between each channel that causes each channel in the last layer of the PINN and BsPINN to specialize in learning different local features, thereby enabling the BsPINN to outperform the PINN.

\subsection{Three-dimensional Helmholtz equation in a 3D M\"{o}bius knot} \label{he3d}
In this subsection, we address the three-dimensional Helmholtz equation within a 3D M\"{o}bius knot, denoting the interior of the model as $\Omega$ and the boundary of the model as $\partial\Omega$. The mathematical expression for the three-dimensional Helmholtz equation is presented as follows
\begin{flalign*}
    &\ \ \ \ \ \ \ \ -\Delta u-\kappa^{2} u = f \text { in } \Omega, &\\
    &\ \ \ \ \ \ \ \ u = u^\ast \text { on } \partial\Omega,&
\end{flalign*}
where $\kappa$ represents the wavenumber, and $f$ represents the external force, which is set to be
\begin{flalign*}
    &\ \ \ \ \ \ \ \ f(x,y,z)=2\kappa^2\sin(\kappa x)\sin(\kappa y)\sin(\kappa z), &
\end{flalign*}
and $u^\ast$ is exact solution which is given by
\begin{flalign*}
    &\ \ \ \ \ \ \ \ u^\ast\left(x,y,z\right)=\sin(\kappa x)\sin(\kappa y)\sin(\kappa z).&
\end{flalign*}
In this scenario, we examine the case where $\kappa=\frac{8\pi}{150}$. We choose this relatively small value of kappa due to the extensive scale of the model.

We compare the performance of the ``5*512'' and ``512-32'' architectures corresponding to PINNs and BsPINNs, respectively.
Throughout the neural network architecture, the sine activation function is utilized in all hidden layers, except for the last layer which utilizes a linear activation function.
We randomly sample $N_f$ points in $\Omega$ as training points for the governing equation, and $N_b$ points on $\partial\Omega$ as training points for the boundary condition.
Here, we consider three distinct scenarios: (a) $N_f=20,000$, $N_b=3,000$, (b) $N_f=40,000$, $N_b=4,000$, and (c) $N_f=60,000$, $N_b=5,000$.
Following the conventions established in section \ref{s2}, we set ${\lambda}_B=100$ in the loss function.
Our training protocol involves a total of $5,000$ iterations, commencing with an initial learning rate of $0.001$.

\begin{figure}[h!]
    \centering
    \includegraphics[width=0.3\textwidth]{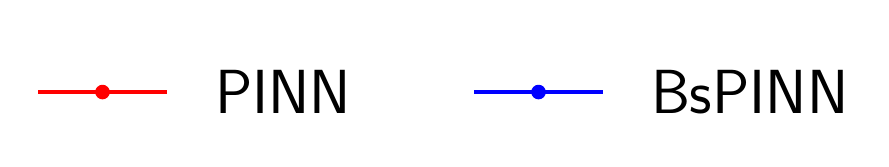} \\
    \begin{subfigure}{0.3\textwidth}
        \includegraphics[width=\textwidth]{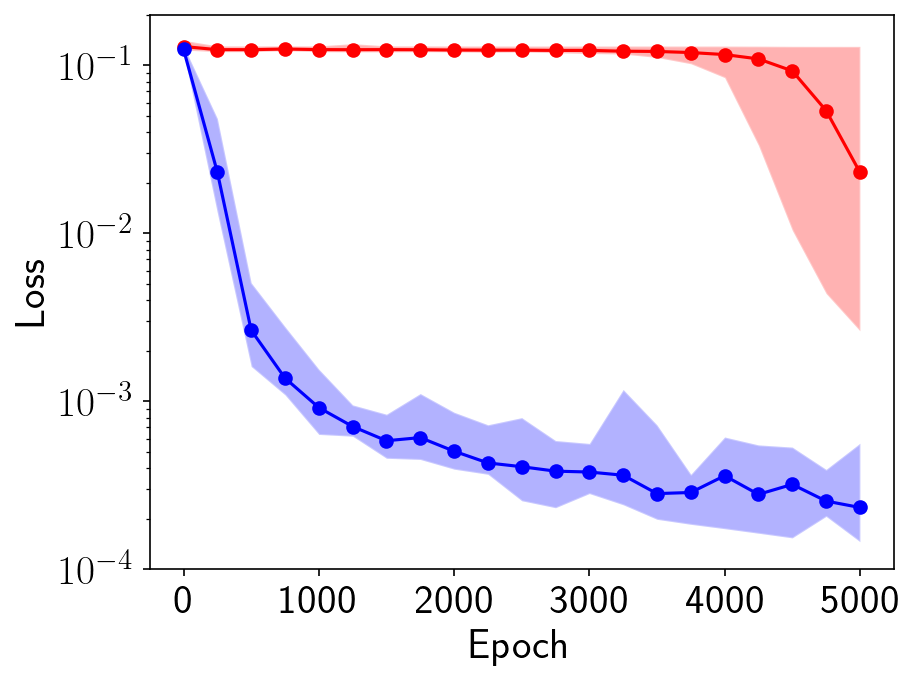}
        \caption{$N_f=20,000$, $N_b=3,000$}
    \end{subfigure}
    \begin{subfigure}{0.3\textwidth}
        \includegraphics[width=\textwidth]{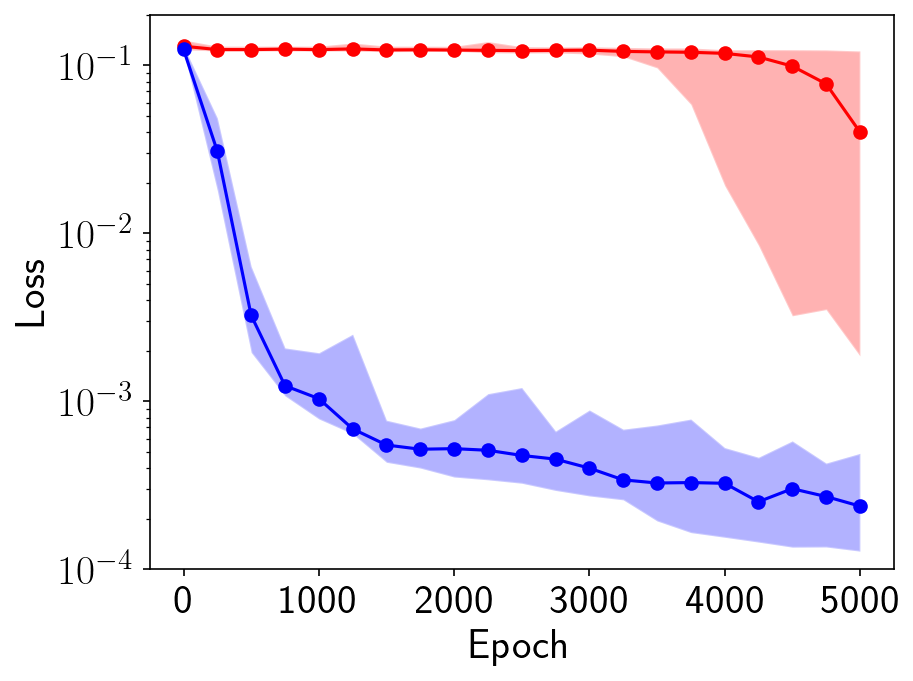}
        \caption{$N_f=40,000$, $N_b=4,000$}
    \end{subfigure}
    \begin{subfigure}{0.3\textwidth}
        \includegraphics[width=\textwidth]{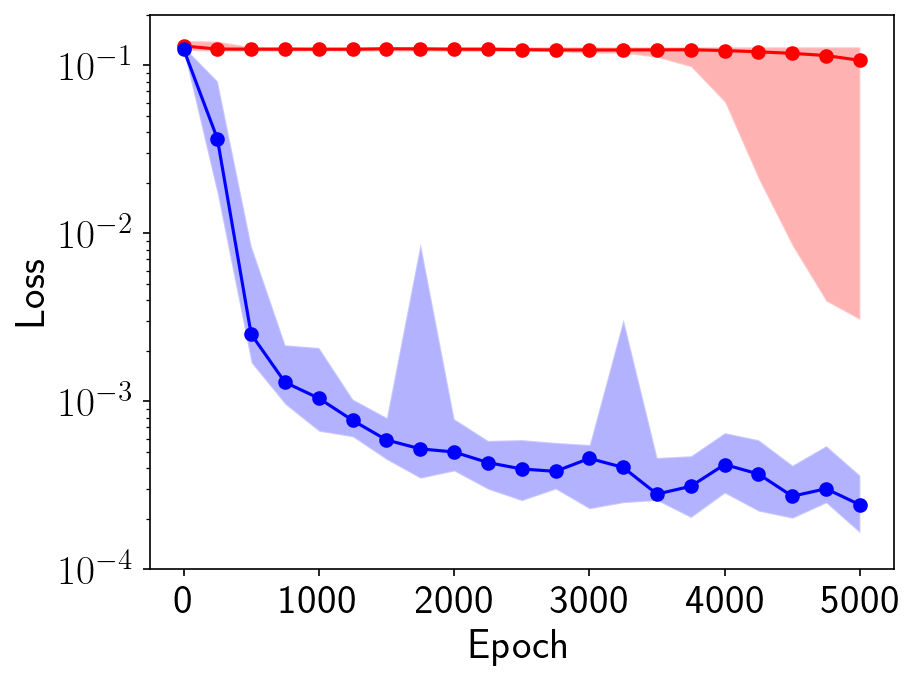}
        \caption{$N_f=60,000$, $N_b=5,000$}
    \end{subfigure}
    \caption{The dynamic change curves of the training loss for the PINNs and BsPINNs when solving the three-dimensional Helmholtz equation in a 3D M\"{o}bius knot. We randomly sample $N_f$ points in $\Omega$ as training points for the governing equation, and $N_b$ points on $\partial\Omega$ as training points for the boundary condition. The shaded regions here represent the segment between the maximum and minimum values of the loss at specific iteration steps from 10 runs with different seeds, while the circular nodes represent the median values.}
    \label{fig:he3d_stl_loss}
\end{figure}

The dynamic change curves of loss values shown in Fig.\ \ref{fig:he3d_stl_loss} illustrate that, in all three scenarios, the BsPINNs experience a swift decline during the early phases of training, converging to a value under $0.001$.
In contrast, the loss values of the PINNs reach a plateau around $0.1$, remaining largely unchanged throughout approximately $4,000$ epochs.
This discrepancy suggests that the BsPINNs, despite sharing identical parameters with the PINNs except for their network structures, exhibit the capability of rapid convergence where the PINNs find it challenging to converge.

When assessing the error, we randomly select $10,000$ points from the interior of the 3D M\"{o}bius knot model and another 10,000 points on the boundary of the 3D M\"{o}bius knot model.
These selected $20,000$ points are denoted as $(x_i,y_i,z_i)(i=1,2,\ldots,N)$ where $N = 2 \times 10^4$.
We calculate the relative error using the formula
\begin{flalign*}
    &\ \ \ \ \ \ \ \ \text{error} = \frac{
            \sqrt{\sum_{i=1}^{N}\left(u_{\theta}\left(x_i,y_i,z_i\right) - u^\ast\left(x_i,y_i,z_i\right)\right)^2}
        }{
            \sqrt{\sum_{i=1}^{N}\left(u^\ast\left(x_i,y_i,z_i\right)\right)^2}
        },&
\end{flalign*}
where $u_{\theta}$ represents the predicted solution computed by neural network parameterized by $\theta$, and $u^\ast$ denotes the reference solution.
The computed relative errors of the PINNs and BsPINNs are shown in Table \ref{tab:error_he3d_stl}.
We observe that in all three scenarios, the relative errors of the BsPINNs are significantly lower than those of the PINNs, and the standard deviations of the BsPINNs' errors are smaller, indicating greater computational stability. %%% deviations用复数对应Table1中有三个标准差
With the increase in the number of training points, the BsPINNs exhibit a slight reduction in relative errors, while the PINNs display fluctuations in their relative errors.
Moreover, considering the model complexities, the PINN architecture entails $1,053,185$ parameters, while the BsPINN structure involves $496,129$ parameters which is approximately $47.1\%$ of the PINN's parameter count.
Overall, the BsPINNs attain enhanced solution accuracy with a smaller number of parameters compared to the PINNs.

\begin{table}[!htbp]
    \centering
    \caption{The relative errors of the PINNs and BsPINNs when solving the three-dimensional Helmhotz equation in a 3D M\"{o}bius knot model using different training points. The results represent mean $\pm$ standard deviation from $10$ runs with diﬀerent random seeds for selecting training points and initializing the neural network parameters.}
        \begin{tabular}{*{4}{c}}
            \hline
            Number of training points & PINNs & BsPINNs \\
            \hline
            $N_f=20,000$, $N_b=3,000$ & $9.554 \times 10^{-1} \pm 4.271 \times 10^{-2}$ & $4.427 \times 10^{-1} \pm 2.811 \times 10^{-2}$ \\
            $N_f=40,000$, $N_b=4,000$ & $9.447 \times 10^{-1} \pm 5.534 \times 10^{-2}$ & $4.350 \times 10^{-1} \pm 4.428 \times 10^{-2}$ \\
            $N_f=60,000$, $N_b=5,000$ & $9.671 \times 10^{-1} \pm 4.820 \times 10^{-2}$ & $4.120 \times 10^{-1} \pm 3.616 \times 10^{-2}$ \\
            \hline
        \end{tabular}
    \label{tab:error_he3d_stl}
    
\end{table}

\begin{figure}[h!]
    \centering
    \begin{subfigure}{.3\textwidth}
        \includegraphics[width=\textwidth, height=.84\textwidth]{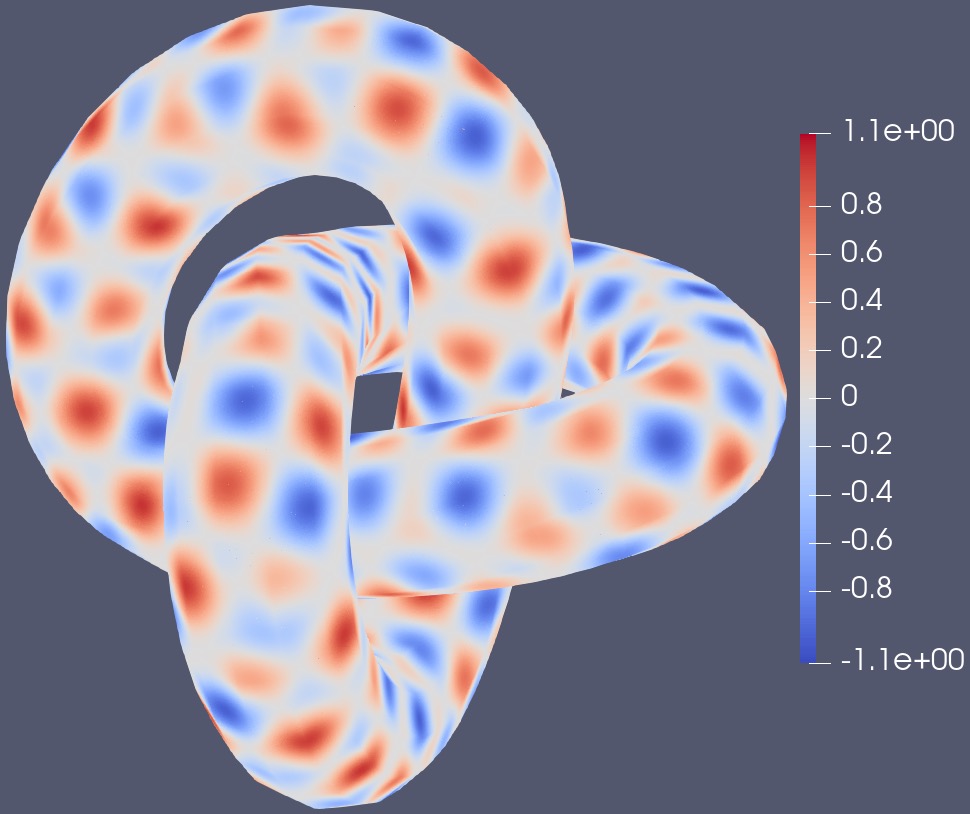}
        \caption{$u$(truth)}
    \end{subfigure}\quad
    \begin{subfigure}{.3\textwidth}
        \includegraphics[width=\textwidth, height=.84\textwidth]{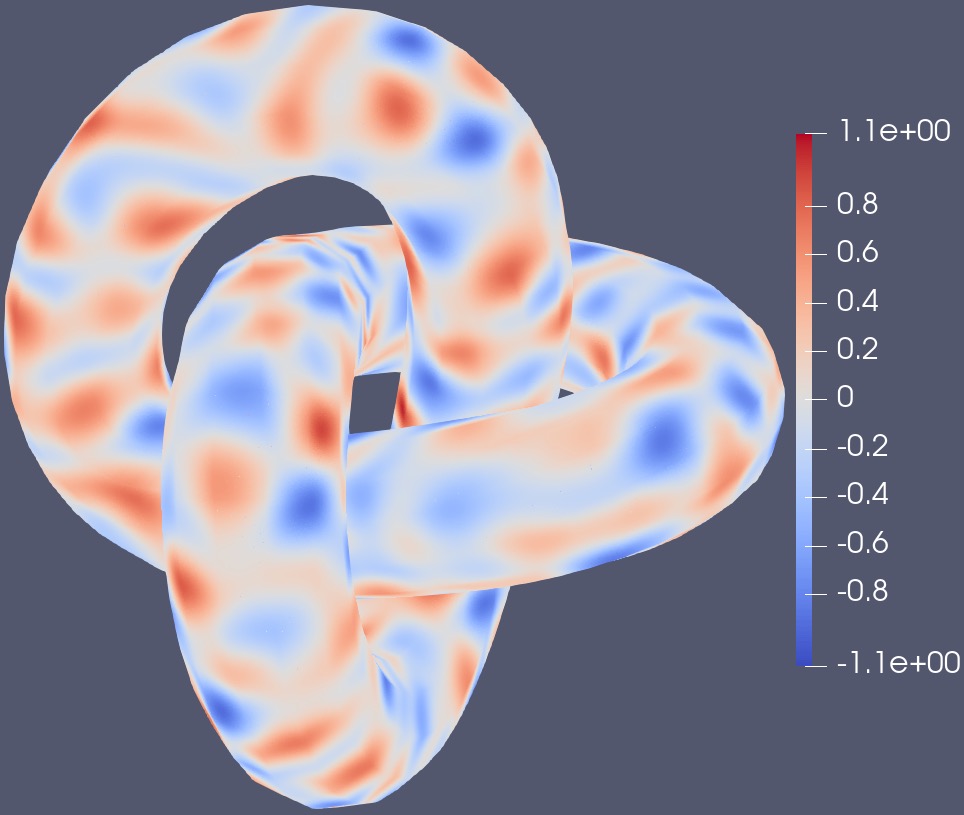}
        \caption{$u$(PINN)}
    \end{subfigure} \\
    \begin{subfigure}{.3\textwidth}
        \includegraphics[width=\textwidth, height=.84\textwidth]{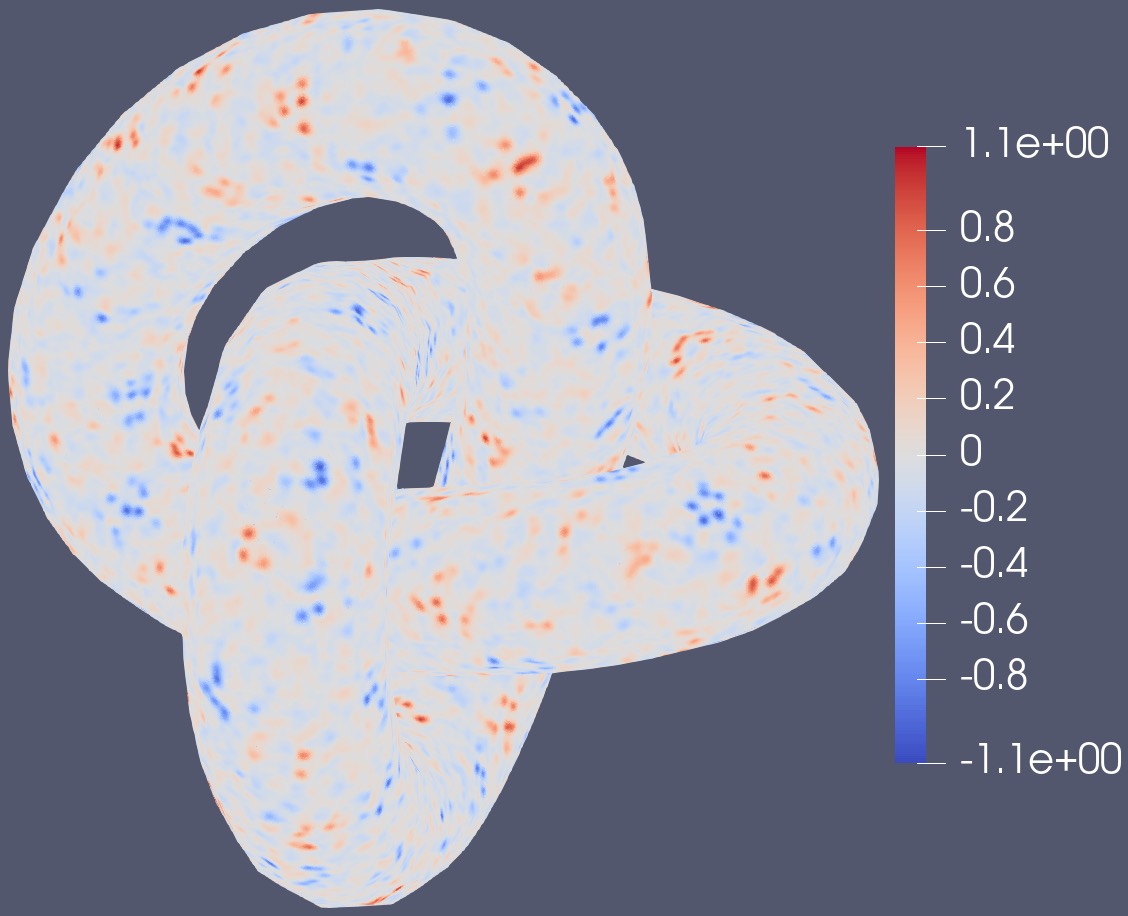}
        \caption{$u$(Modulus)}
    \end{subfigure}\quad
    \begin{subfigure}{.3\textwidth}
        \includegraphics[width=\textwidth, height=.84\textwidth]{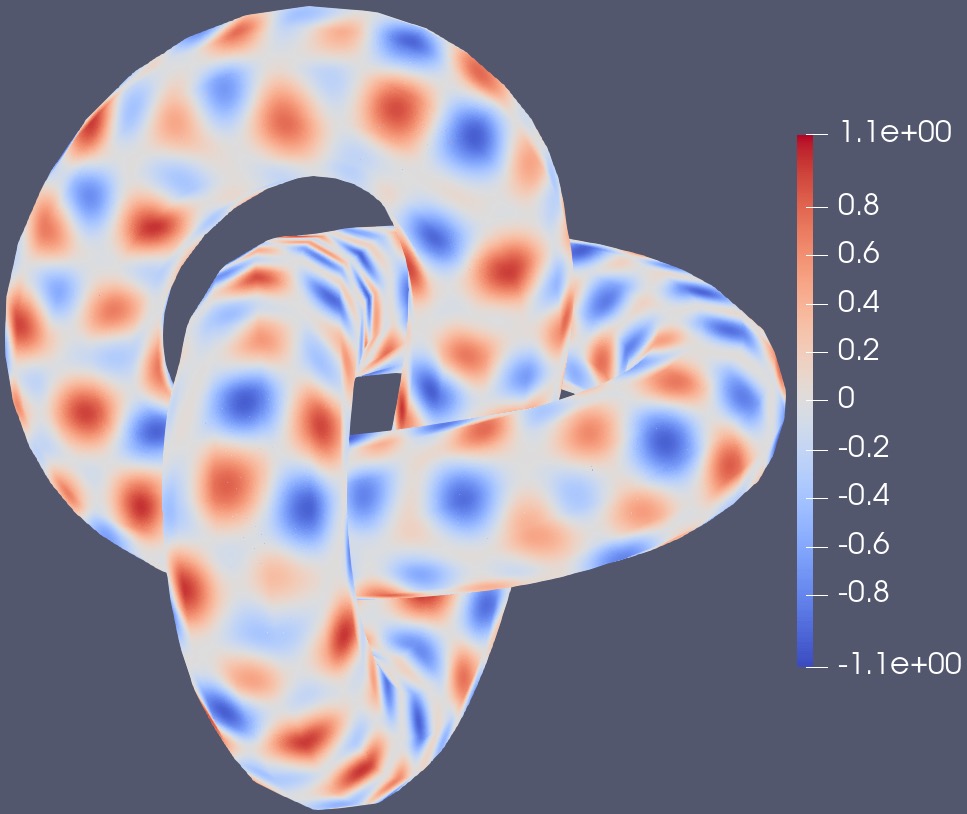}
        \caption{$u$(BsPINN)}
    \end{subfigure}
    \caption{Exact solution (a) and the predicted solutions of the PINN (b), Modulus (c) and the BsPINN (d) in terms of the displacement value $u$ for the three-dimensional Helmholtz equation in a 3D M\"{o}bius knot model.}
    \label{fig:he3d_stl_u}
\end{figure}

From Fig.\ \ref{fig:he3d_stl_u} where we present the images of exact and predicted solutions on $\partial\Omega$, we can observe that the BsPINN's predicted solution aligns more closely with the exact solution, while the PINN fails to capture the intricate details of oscillations.
For instance, in Fig.\ \ref{fig:he3d_stl_u}(b), some peaks and valleys are not separated but instead connected together.
These results conclusively underscore the advantage of BsPINNs in solving PDEs characterized by oscillations.

For further comparison, we employed the existing commercial software, NVIDIS Modulus \cite{nvidia_modulus}, to solve this equation. We use $N_f=40,000$ and $N_b=4,000$ for the training points, as this configuration minimizes the average relative error of PINNs. As Modulus solely supports exponential decay of learning rate, we implemente an exponential decay of $0.9$ every $1,000$ epochs during training.
All other hyper-parameters of Modulus remain consistent with the PINNs used in this experiment.
As shown in Fig.\ \ref{fig:he3d_stl_u}(c), the solution obtained by Modulus exhibits substantial errors, failing to accurately fit the peaks and valleys.

\subsection{High-dimensional oscillatory Poisson equation} \label{s_pos}
High-dimensional PDEs, including those with four or more dimensions, present significant challenges for conventional numerical methods such as FEM, primarily because of the need to partition intricate high-dimensional grids.
In this context, we employ the residual neural network, as introduced in \cite{zeng2022adaptive}, to address the resolution of the high-dimensional Poisson equation.

Consider the high-dimensional Poisson equation
\begin{flalign*}
    &\ \ \ \ \ \ \ \ - \Delta u=f\ \text{in}\ \Omega, &\\
    &\ \ \ \ \ \ \ \ u = g\ \text{on}\ \partial\Omega, &
\end{flalign*}
where $\Omega=(-1,1)^d$,
\begin{flalign*}
    &\ \ \ \ \ \ \ \ f(x)=d \cdot c^2 \cdot \sin\left(c\sum_{i=1}^{d}x_i\right)-\frac{2}{d}, & \\
    &\ \ \ \ \ \ \ \ g(x)=\left(\frac{1}{d}\sum_{i=1}^{d}x_i\right)^2+\sin\left(c\sum_{i=1}^{d}x_i\right).&
\end{flalign*}
The exact solution is given by
\begin{flalign*}
    &\ \ \ \ \ \ \ \ u^\ast(x)=\left(\frac{1}{d}\sum_{i=1}^{d}x_i\right)^2+\sin\left(c\sum_{i=1}^{d}x_i\right). &
\end{flalign*}
Here, we set $d=10$ and $c=0.6\pi$.

We compare the performance of the following three sets of residual block structures: (a) ``4*256'' versus ``256-32'', (b) ``5*256'' versus ``256-16'', (c) ``6*256'' versus ``256-8'', corresponding to the PINNs and BsPINNs respectively, with a total of $2$ residual blocks in the residual neural network.
Within each residual block, we utilize the sine activation function in all hidden layers.
For the training points corresponding to the governing equation, we randomly sample $4,000$ points in $\Omega$.
For the training points corresponding to the boundary condition, we sample $200$ points randomly from each edge of $\Omega$.
Here, an edge of $\Omega$ signifies that the element in a specific dimension reach either $1$ or $-1$. This results in a total of $4,000$ training points for the boundary conditions.
Following the notation outlined in section \ref{s2}, we set ${\lambda}_B=1$ in the loss function.
For the training process, we fix the total number of iterations to $10,000$ and initialize the learning rate to $0.001$.

\begin{figure}[h!]
    \centering
    \includegraphics[width=0.3\textwidth, height=0.05\textwidth]{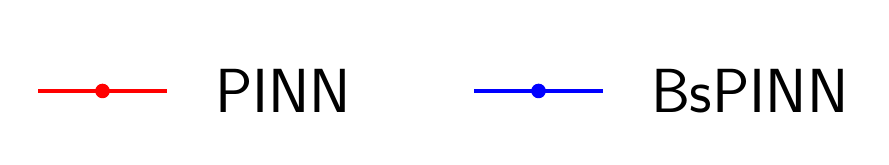} \\
    \begin{subfigure}{0.3\textwidth}
        \includegraphics[width=\textwidth]{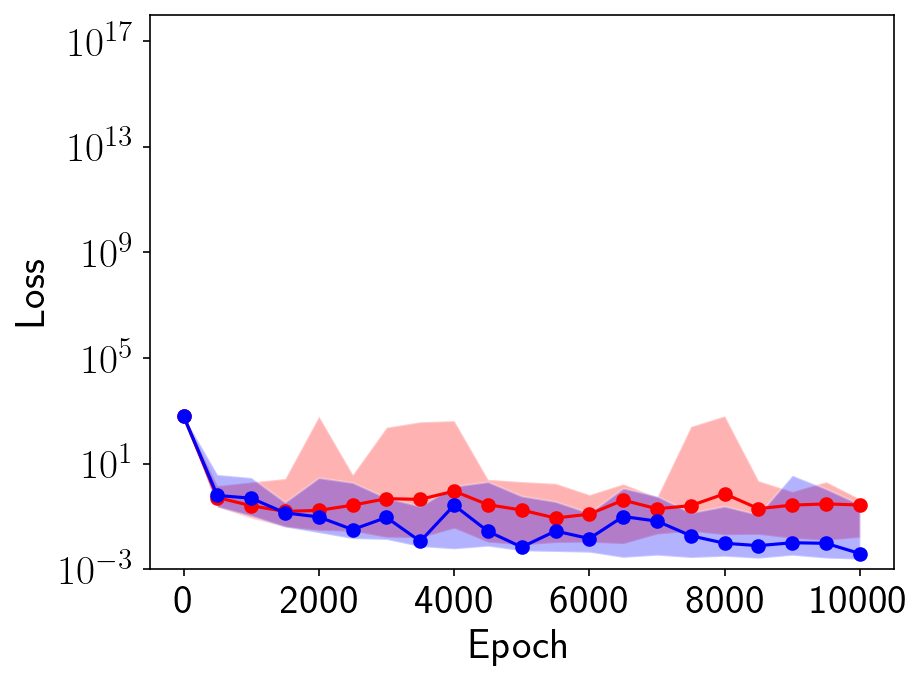}
        \caption{``4*256'' versus ``256-32''}
    \end{subfigure}
    \begin{subfigure}{0.3\textwidth}
        \includegraphics[width=\textwidth]{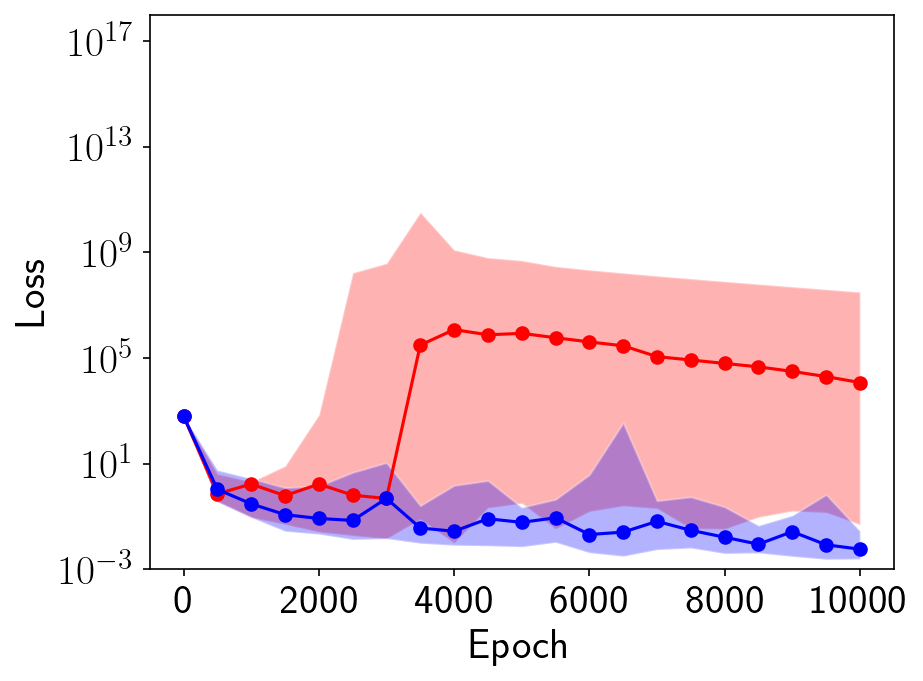}
        \caption{``5*256'' versus ``256-16''}
    \end{subfigure}
    \begin{subfigure}{0.3\textwidth}
        \includegraphics[width=\textwidth]{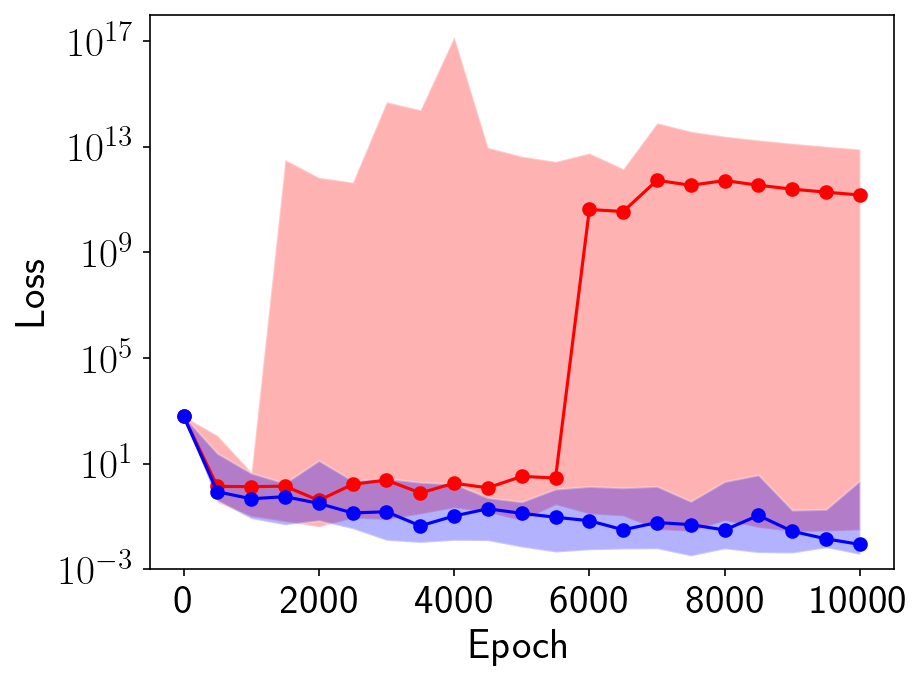}
        \caption{``6*256'' versus ``256-8''}
    \end{subfigure}
    \caption{Variations in the loss values for the PINNs and BsPINNs during the training process for solving the high-dimensional oscillatory Poisson equation. The shaded regions here represent the segment between the maximum and minimum values of the loss at specific iteration steps from $10$ runs with different seeds, while the circular nodes represent the median values.}
    \label{fig:highpos_loss}
\end{figure}

\begin{figure}[h!]
    \centering
    \begin{subfigure}{0.3\textwidth}
        \includegraphics[width=\textwidth]{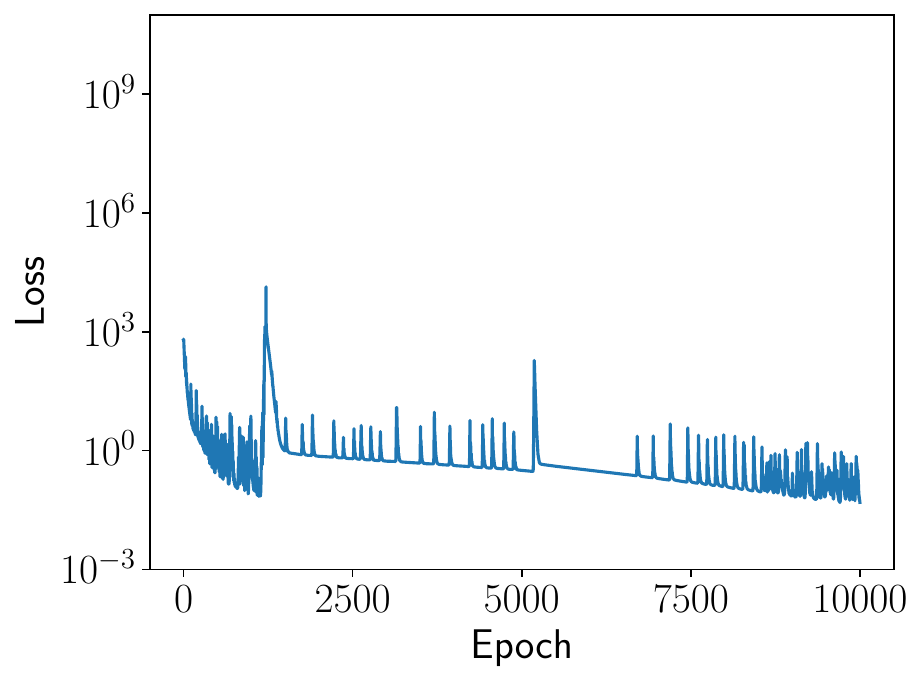}
        \caption{}
    \end{subfigure}
    \begin{subfigure}{0.3\textwidth}
        \includegraphics[width=\textwidth]{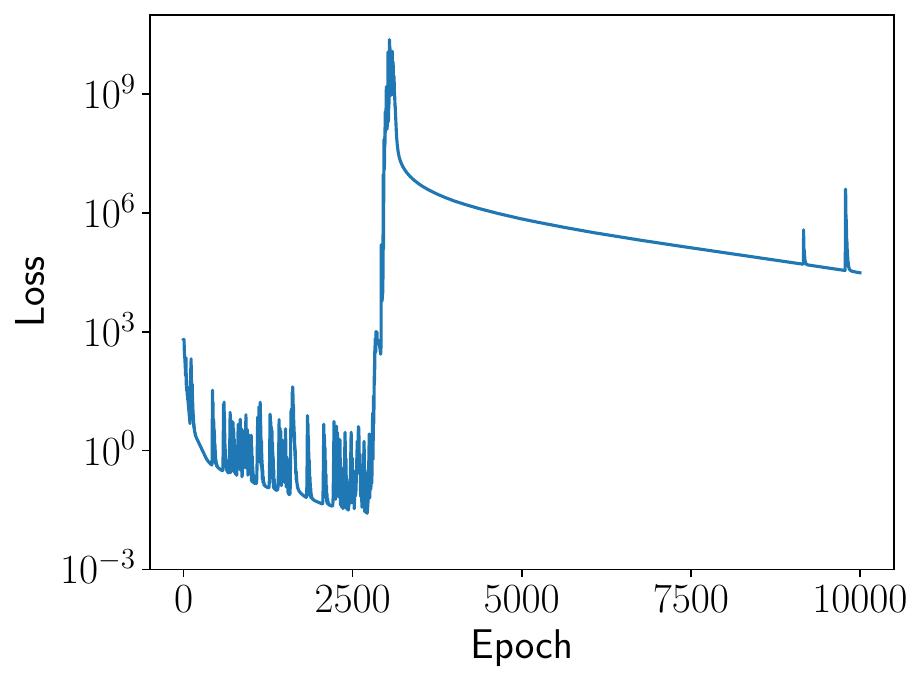}
        \caption{}
    \end{subfigure}
    \caption{Two distinct dynamic trends in the loss values during the training of the PINNs with structure ``5*256''. That is, (a) the loss value consistently does not increase to a substantial level, and (b) the loss value abruptly increases to a substantial level and does not decrease back to the same order of magnitude as before the increase.}
    \label{fig:loss_poisson_increase}
\end{figure}

Depicted in Fig.\ \ref{fig:highpos_loss} are the dynamic changes of the loss value with respect to the training steps.
In this experiment, we observe that during the training process, the PINNs' loss values exhibit two distinct trends: (a) the loss value consistently does not increase to a substantial level, and (b) the loss value abruptly increases to a substantial level and does not decrease back to the same order of magnitude as before the increase, which are illustrated in Fig.\ \ref{fig:loss_poisson_increase}(a) and (b) respectively.
We also observe in this experiment that the loss values of the PINNs exhibit a sudden increase in $7$ out of $10$ runs when there are five hidden layers, and $8$ out of $10$ runs when there are six hidden layers.
From Fig.\ \ref{fig:highpos_loss}, we can see that, when the neural network has a large number of hidden layers, the PINNs' loss values tend to exhibit a sudden and substantial increase, and the greater the number of hidden layers, the higher the values reach after the abrupt increase. After the sudden increase in the loss value, the loss values of the PINNs stagnate at around $10^{5}$ in the case of (b) and $10^{13}$ in the case of (c), and do not decrease back to the same order of magnitude as before the increase, which implies that the training of the PINNs after the sharp increase in loss values is ineffective. Fortunately, we do not observe this phenomenon when tracing the BsPINNs. And from Table \ref{tab:min_loss_poisson}, it is evident that in all three scenarios, both the minimum training loss values and the standard deviations for the BsPINNs are lower than those of the PINNs, demonstrating the convergence and stability of BsPINNs. %%% 这里用values和deviations是对应Table2中包含的3个最小训练损失和标准差。

This phenomenon bears a striking resemblance to the over-smoothing observed in graph neural networks (GNNs) \cite{2020graph}, which describes a significant performance degrade when stacking more layers. To address the over-smoothing in GNNs, researchers proposed the DropEdge algorithm \cite{rong2020dropedge}, which randomly removes a certain number of edges from the input graph at each training epoch.
Interestingly, in BsPINNs, by reducing certain connections between neurons before training the neural network, the issue of ``over-smoothing" encountered with an increase in the number of hidden layers in PINNs is effectively mitigated, achieving a result similar to the DropEdge method.

\begin{table}[!htbp]
    \centering
    \caption{The minmum training loss values of the PINNs and BsPINNs when solving the high-dimensional Poisson equation using different network structures. The results represent mean $\pm$ standard deviation from 10 runs with diﬀerent random seeds for selecting training points and initializing the neural network parameters.}
        \begin{tabular}{*{4}{c}}
            \hline
            Network structures & PINNs & BsPINNs \\
            \hline
            ``4*256'' versus ``256-32'' & $8.831 \times 10^{-3} \pm 4.968 \times 10^{-3}$ & $3.206 \times 10^{-3} \pm 2.276 \times 10^{-3}$ \\
            ``5*256'' versus ``256-16'' & $2.376 \times 10^{-2} \pm 1.653 \times 10^{-2}$ & $3.221 \times 10^{-3}  \pm 1.086 \times 10^{-3}$ \\
            ``6*256'' versus ``256-8'' & $2.979 \times 10^{-2} \pm 2.333 \times 10^{-2}$ & $3.320 \times 10^{-3} \pm 1.257 \times 10^{-3}$ \\
            \hline
        \end{tabular}
    \label{tab:min_loss_poisson}
\end{table}

\begin{table}[!htbp]
    \centering
        
        \caption{The relative errors of the PINNs and BsPINNs when solving the high-dimensional Poisson equation using different network structures. The results represent mean $\pm$ standard deviation from $10$ runs with diﬀerent random seeds for selecting training points and initializing the neural network parameters.}
        \begin{tabular}{*{4}{c}}
            \hline
            Network structures & PINNs & BsPINNs \\
            \hline
            ``4*256'' versus ``256-32'' & $3.130 \times 10^{-4} \pm 2.706 \times 10^{-4}$ & $1.112 \times 10^{-4} \pm 2.372 \times 10^{-5}$ \\
            ``5*256'' versus ``256-16'' & $3.941 \times 10^{-4} \pm 1.486 \times 10^{-4}$ & $1.866 \times 10^{-4}  \pm 9.842 \times 10^{-5}$ \\
            ``6*256'' versus ``256-8'' & $6.630 \times 10^{-4} \pm 4.282 \times 10^{-4}$ & $3.367 \times 10^{-4} \pm 1.329 \times 10^{-4}$ \\
            \hline
        \end{tabular}
    \label{tab:error_poisson}
\end{table}

For evaluating the error, the formulation for calculating relative error is
\begin{flalign*}
    &\ \ \ \ \ \ \ \ \text{error} = \frac{
            \int_{\Omega}|u_{\theta} - u^\ast|^2
        }{
            \int_{\Omega}|u^\ast|^2
        },&
\end{flalign*}
where $u_{\theta}$ represents the predicted solution computed by neural network parameterized by $\theta$, and $u^\ast$ denotes the reference solution.
To compute the integral in ten-dimensional space, we extended the one-dimensional four-point Gaussian quadrature formula \cite{burden2015numerical} to ten dimensions.
This means that there are a total of $4^{10}$ ten-dimensional integration points, and the weight for each integration point is the product of one-dimensional weights corresponding to elements in each dimension.
As illustrated in Table \ref{tab:error_poisson}, the computed relative errors for both the PINNs and BsPINNs are $O(10^{-4})$, with the BsPINNs exhibiting lower relative errors compared to the PINNs.
Simultaneously, in all three scenarios, the standard deviations of the relative errors for the BsPINNs are significantly lower than those of the PINNs, demonstrating the stability of the BsPINNs' performance.

\section{Conclusion} \label{s5}
This paper introduces Binary Physics-Informed Neural Networks (BsPINNs), a novel class of PINNs based on BsNNs.
By solving various PDEs characterized by rapidly-changing solutions, it is demonstrated that BsPINNs surpasse the performance of PINNs in terms of convergence speed, especially at early training, as well as the solution accuracy.
Moreover, to eliminate randomness, we compare the performance of BsPINNs and PINNs under different network architectures and varying numbers of training points, confirming the stability of BsPINNs' performance.
Upon further analysis of the experimental results, we find that BsPINNs can capture the local features of rapidly changing solutions and address the ``over-smoothing" issue encountered by PINNs, as well as the problem arising from the reduced proportion of training points for the ``tail category".
Our proposed method offers a promising and straightforward approach to enhance the efficiency, robustness, and accuracy of neural network-based approximations for nonlinear functions, as well as for solving partial differential equations.

However, there remain several aspects of this neural network architecture that require further investigation.
For instance, despite the potential training speed advantage indicated by the parameter count discrepancy between PINN and BsPINN, this advantage might not be fully realized.
It appears that existing neural network function libraries might not effectively leverage BsPINN's advantages stemming from its reduced parameter count.
Also, there is a need to establish mathematical proofs demonstrating how this type of neural networks effectively enhances PINNs' performance in tackling PDEs with highly rapidly-changing solutions.
Additionally, the suitability of the commonly used optimizer, Adam, which is prevalent in image classification problems, for training BsPINNs could be explored. Identifying optimizers better tailored to this problem might be a promising avenue for future research.\\

\noindent{\small\textbf{Funding}
This work is partially
supported by the National Key R\&D Program of China (2022ZD0117805), the Key-Area Research and Development Program of Guangdong Province (No.2021B 0101190003), and the Natural Science Foundation of Guangdong Province, China (No.2022A1515010831).}

\noindent{\small \textbf{Data Availability}
The datasets generated during the current study are available from the corresponding author
upon reasonable request.}

\section*{Declarations}
\noindent{\small\textbf{Competing Interests} The authors declare that they have no conflict of interest.}

\bibliographystyle{plain}
\bibliography{reference}

\end{document}